\documentclass[10pt,twocolumn,letterpaper]{article}

\usepackage{cvpr}              

\definecolor{cvprblue}{rgb}{0.21,0.49,0.74}
\usepackage[pagebackref,breaklinks,colorlinks,allcolors=cvprblue]{hyperref}
\usepackage{multirow}
\definecolor{lightblue}{rgb}{0.9, 0.95, 1}
\usepackage{listings}
\usepackage{tcolorbox}
\usepackage{mdframed}

\def\etc{\emph{etc}.~}
\def\benchname{MotionBench\xspace}
\definecolor{dt}{gray}{0.6}
\definecolor{dtdark}{gray}{0.5}

\newcommand{\Tref}[1]{Table~\ref{#1}}

\newcommand{\Fref}[1]{Figure~\ref{#1}}

\newcommand{\fref}[1]{Fig.~\ref{#1}}
\newcommand{\sref}[1]{Sec.~\ref{#1}}

\newtcolorbox{promptbox}[1][]{
  breakable,
  title=#1,
  colback=gray!5,
  colframe=black,
  colbacktitle=gray!15,
  coltitle=black,
  bottomrule=1.5pt,
  toprule=1.5pt,
  leftrule=1pt,
  rightrule=1pt,
  arc=0pt,
  outer arc=0pt,
  enhanced,
}

\def\paperID{xxxx} 
\def\confName{CVPR}
\def\confYear{2025}

\title{MotionBench: Benchmarking and Improving Fine-grained Video Motion Understanding for Vision Language Models 
}

\author{Wenyi Hong\textsuperscript{1}\thanks{Equal contribution.} \ \ Yean Cheng\textsuperscript{2}\footnotemark[1] \ \ Zhuoyi Yang\textsuperscript{1}\footnotemark[1] \ \ Weihan Wang\textsuperscript{2}\ \ Lefan Wang\textsuperscript{2} \ \ \\ Xiaotao Gu\textsuperscript{2} \ \  Shiyu Huang\textsuperscript{2} \ \ 
Yuxiao Dong\textsuperscript{1}\footnotemark[2]  \ \ Jie Tang\textsuperscript{1}\thanks{Corresponding authors}
\\
  \textsuperscript{1}Tsinghua University\ \ \textsuperscript{2}Zhipu AI\ \ 
  \\
  \texttt{wenyi.hong@outlook.com, cya17@tsinghua.org.cn,}  \\ \texttt{zhuoyiyang2000@gmail.com, jietang@tsinghua.edu.cn} \\
}

\begin{document}

\maketitle

\makeatletter
{\let\thefootnote\relax\footnotetext{Work was done when WH, ZY interned at Zhipu AI.}}
\makeatother

\begin{abstract}

In recent years, vision language models (VLMs) have made significant advancements in video understanding. However, a crucial capability — fine-grained motion comprehension — remains under-explored in current benchmarks. To address this gap, we propose MotionBench, a comprehensive evaluation benchmark designed to assess the fine-grained motion comprehension of video understanding models. 
MotionBench evaluates models' motion-level perception through six primary categories of motion-oriented question types and includes data collected from diverse sources, ensuring a broad representation of real-world video content.
Experimental results reveal that existing VLMs perform poorly in understanding fine-grained motions.
To enhance VLM's ability to perceive fine-grained motion within a limited sequence length of LLM, we conduct extensive experiments reviewing VLM architectures optimized for video feature compression and propose a novel and efficient Through-Encoder (TE) Fusion method.
Experiments show that higher frame rate inputs and TE Fusion yield improvements in motion understanding, yet there is still substantial room for enhancement.
Our benchmark aims to guide and motivate the development of more capable video understanding models, emphasizing the importance of fine-grained motion comprehension. Project page: \url{https://motion-bench.github.io}.

\end{abstract}    
\section{Introduction}

\begin{figure}[t]
    \centering
    \includegraphics[width=0.99\linewidth]{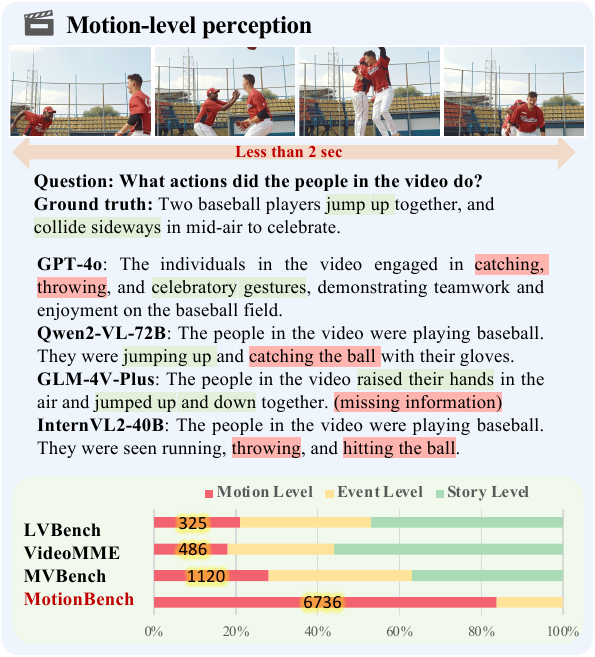}
    \caption{State-of-the-art video understanding models struggle with basic motion-level perception. Compared to existing benchmarks, our proposed \benchname focuses on assessing the model's Motion level perception capability, which is critical in understanding videos with fast and instant interactions and motions.}
    \vspace{-1em}
    \label{fig:teaser}
\end{figure}

With the rapid development of pre-training, an increasing number of studies focus on leveraging large vision language models (VLMs) for video understanding~\cite{reid2024gemini, liu2025st,hong2024cogvlm2,liu2024oryx,li2024aria}.
For instance, CogVLM2-Video~\citep{hong2024cogvlm2}, LLaVA-Video~\citep{zhang2024video} and PLLaVA~\citep{xu2024pllava} continually train image-understanding models to achieve video-understanding models, and Qwen2-VL\citep{wang2024qwen2}, LLaVA-OneVision~\citep{li2024llava} explore mixed training upon both images and videos. 
To effectively evaluate video understanding VLMs as well as guide further advancement, a series of video understanding benchmarks emerged, with focuses on general video understanding capability~\citep{li2024mvbench, li2024videovista, wang2024lvbench,fu2024video} or specific capabilities such as long video understanding~\citep{wang2024lvbench, zhou2024mlvu, wu2024longvideobench}. 
Video understanding questions can be categorized into three levels based on the granularity of understanding: \emph{motion-level} (capturing fine-grained motion), \emph{event-level} (addresses distinct segments of activities~\citep{du2024towards}), and \emph{story-level} (a holistic understanding of the storyline across the video~\citep{ghermi2024short}).
Among them, motion-level understanding acts as a foundational ability and plays a pivotal role in applications such as anomaly detection, open-domain action analysis, detailed video captioning, \etc 
However, while some benchmarks shifted their focus toward \emph{event-} and \emph{story-level} understanding, most benchmarks lack a dedicated set for evaluating \emph{motion-level} understanding. 
To quantitatively analyze the granularity distribution across benchmarks, we leverage GPT-4o\footnote{gpt-4o-2024-08-06} for question analysis. The results in Figure~\ref{fig:teaser} indicate that the foundational motion-level comprehension is being overlooked, with the data volume and diversity for \emph{motion-level} content being limited. 
Some datasets from earlier years focused on \emph{low-level} action recognition within specific domains, but their content and categories are highly constrained.

Is this because \emph{motion-level} understanding is too trivial to merit attention? 
To answer this question, we build \benchname to thoroughly evaluate the \emph{motion-level} capability of current video models. \benchname comprises 8,052 questions covering six main categories of video motion, with diverse video collected from the web (Panda-70M~\cite{chen2024panda}, Pexels\footnote{\url{https://www.pexels.com}}), public datasets (MedVid~\cite{gupta2023dataset}, SportsSloMo~\cite{chen2024sportsslomo}, Ha-ViD~\cite{zheng2024ha}), and self-synthetic videos generated via Unity\footnote{\url{https://unity.com/cn}}, capturing a broad distribution of real-world application. Surprisingly, most state-of-the-art models can only achieve accuracy lower than 60\%, significantly below the threshold for practical applications, which highlights two primary technical challenges:

\noindent{\bf High Frame Rate vs. Computational Cost:} The first challenge lies in the contradiction between the high frame rate required for fine-grained motion understanding and the high computational cost of long sequence lengths.
Long sequence lengths substantially increase the computational and memory burden in both training and inference. Consequently, most current video understanding models can only handle a limited number of frames, falling short of the demands for fine-grained motion analysis.
For example, Intern-VL2~\cite{chen2023internvl}, LLaVA-Next-Video~\cite{zhang2024llavanextvideo} and CogVLM2-Video~\cite{hong2024cogvlm2} can only accept 16 to 64 frames, thus can only sample frames at an extreme-low rate of 1 frame every 5 seconds (\ie, 0.2 fps) for a 5-minute video which is common in daily life. 
To address this, we \textbf{conduct the first comprehensive evaluation} over existing video feature compression architectures and identify their common shortcomings-shallow fusion. Based on these findings, \textbf{we propose a novel VLM architectural paradigm---Through-Encoder Fusion} (TE Fusion), which enhances video feature representation under a fixed decoder sequence length by applying deep fusion throughout the visual encoder. Experiments on benchmarks across various video lengths and contents demonstrate that TE Fusion achieves state-of-the-art performance, and shows particular advantages under high compression ratios.

\noindent{\bf Limited Fine-Grained Motion Understanding:} The second challenge arises from the limited foundational capability to comprehend fine-grained motion in current video understanding models. While a higher frame rate brings some performance improvements (\cref{tab:compress_normal_main}), models' \emph{motion-level} understanding remains constrained, achieving accuracies of below 60\% on \benchname (\cref{tab:dataset eval}). To address this, \textbf{we additionally release a dataset of 5,000 videos with manually annotated fine-grained motion descriptions}, which are annotated and double-checked together with the benchmark annotation process (refer to \cref{fig:dy info anno} for example). Each video includes dynamic information descriptions with annotation density reaching 12.63 words per second, providing researchers with resources for further development and training to enhance video models’ \emph{motion-level} comprehension capabilities.

\noindent{\bf Contribution.} Our main contributions include:
\begin{itemize}
    \item We introduce \benchname, the largest \emph{motion-level} video benchmark, featuring a wide range of video sources and question types, along with a carefully designed annotation pipeline that ensures diversity and accuracy.
    \item \benchname reveals a critical deficiency in \emph{motion-level} understanding among current video understanding models, which is largely overlooked by existing research.
    \item We propose TE Fusion, a novel compression architecture to enhance \emph{motion-level} understanding under constrained LLM context length. Experimental results demonstrate that TE Fusion achieves state-of-the-art results on MotionBench and outperforms other compression methods across \benchname, MVBench~\cite{li2024mvbench}, LVBench~\cite{wang2024lvbench}, and VideoMME~\cite{fu2024video} in the ablation study, and shows a particular advantage in high compression ratio scenarios. 
\end{itemize}








\section{Related Work}

\begin{table*}[ht]
\centering
\small
\caption{The comparison of existing video VLM benchmarks with MotionBench. MotionBench collects various video sources including web videos and synthetic videos, and provides a new evaluation perspective in motion level perception.}\label{tab:related-work}
\resizebox{\textwidth}{!}{%
\setlength{\tabcolsep}{3mm}{
\begin{tabular}{cccccc}
\toprule
Benchmarks                        & \#Videos & \#QAs & Perception Level              & Data source               & Dataset Feature             \\ \hline
MVBench~\cite{li2024mvbench}                           & 4,000    & 4,000 & general, motion\textless 30\%  & existing datasets                             & general \\
TempCompass~\cite{liu2024tempcompass}                       & 410      & 1,580 & general, motion\textless 20\%  & ShutterStock                          & temporal concept      \\
VideoMME~\cite{fu2024video}                          & 900      & 2,700 & general, motion\textless 20\% & Youtube                                      & general \\
AutoEval-Video~\cite{chen2023autoevalvideo}                    & 327      & 327   & event level                   & Youtube                                      & open-ended QA               \\
EgoSchema~\cite{mangalam2023egoschema}              & 5,031    & 5031  & event level                   & ego-centric video                     & ego-centric           \\
LVBench~\cite{wang2024lvbench}                           & 103      & 1,549 & event \& story level          & Youtube                                      & long video    \\
LongVideoBench~\cite{wu2024longvideobench}                    & 3,763    & 6,678 & event \& story level          & web channels                                 & long videos    \\
MovieChat-1K~\cite{song2024moviechat}                      & 130      & 1,950 & story level                   & movies                                        & movie         \\
Short Film Dataset~\cite{ghermi2024short}    & 1,078    & 4,885 & story level                   & short films                           & story-level    \\ \hline
MotionBench                       &      5,385    &   8,052   & motion level                  & \begin{tabular}[c]{@{}c@{}} web videos, movies, \\ synthetic videos, datasets\end{tabular}                     & motion perception  \\ \bottomrule
\end{tabular}
}}
\vspace{-1em}
\end{table*}

\subsection{Video Understanding Benchmarks}
To effectively evaluate video understanding models and drive their advancement, a series of benchmarks are proposed.
Traditional benchmarks like MSRVTT-QA~\citep{xu2016msr} and ActivityNet-QA~\citep{yu2019activitynet} primarily focus on basic action recognition and video question answering with short clips. While these benchmarks provide a foundation for assessing video understanding capabilities, they lack the granularity to evaluate subtle motion comprehension.
Recently, more benchmarks emerged to assess video VLMs, as shown in \cref{tab:related-work}. MVBench~\citep{li2024mvbench} emphasizes general video understanding, introducing 20 temporal-related tasks across six domains. Video-MME~\cite {fu2024video} offers an evaluation framework featuring videos of varying durations—from 11 seconds to over an hour—while incorporating multimodal elements such as subtitles and audio. Some benchmarks focus on specific, challenging capabilities. For example, LVBench~\cite{wang2024lvbench}, LongVideoBench~\citep{wu2024longvideobench}, and MLVU~\citep{zhou2024mlvu} target event- or story-level understanding across long temporal horizons. 
However, these benchmarks primarily focus on general video understanding, lacking a dedicated dataset or subset specifically designed for motion-level assessment. This limitation results in reduced volume and diversity in evaluating motion dynamics. Furthermore, most benchmarks rely on data from a single source, falling short of representing a comprehensive distribution of downstream applications.

To address these gaps, we propose MotionBench, a benchmark dedicated to fine-grained motion understanding. By leveraging data from seven distinct sources and encompassing six motion-oriented task categories, MotionBench offers a diverse range of video content and a specialized focus on motion-level perception, advancing the evaluation of video understanding models in this crucial area.

\subsection{VLMs for video understanding}
Recent advancements in Visual Language Models (VLMs) have demonstrated significant potential in video understanding, mostly extending pre-trained VLMs~\citep{liu2024visual, wang2023cogvlm} to handle video modality. 
Video VLMs typically comprise three core components: a visual encoder for visual feature extraction, a modality alignment module to integrate visual features into the language model's embedding space, and an LLM backbone for decoding multi-modal context. A straightforward architecture is LLaVA-Next-Video~\citep{zhang2024llavanextvideo}, CogVLM2-Video~\citep{hong2024cogvlm2} and Intern-VL2~\cite{internvl2}, where videos are treated as sequences of images, extending VLM's strong image understanding capabilities to videos. Qwen2-VL~\citep{Qwen2VL} further introduces 3D-RoPE to enable understanding of arbitrary-length videos. 
However, the high computational and memory demands of handling high-frame-rate, long-duration videos have prompted initial explorations into video compression in both pixel and feature spaces.
For instance, InternVideo2~\citep{wang2024internvideo2} and Video-LLaMA~\citep{zhang2023videollama} adopt QFormer~\citep{li2023blip} for video feature extraction, PLLaVA~\citep{xu2024pllava} utilizes adaptive pooling, Kangaroo~\citep{liu2024kangaroo} employs a unified spatial-temporal patchification, and Qwen2-VL~\citep{Qwen2VL} fuses neighboring frames before visual encoder. 

Despite these advancements, to our knowledge, no comprehensive and fair comparison exists among these compression methods and evaluating their performance as compression ratios increase. Moreover, current approaches are generally limited to shallow fusion that is confined to the compression operator itself, which restricts their performance, especially in high compression rate scenarios

\section{MotionBench: Motion-Level Benchmarking}
 \begin{figure*}[t]
    \centering
    \includegraphics[width=0.99\linewidth]{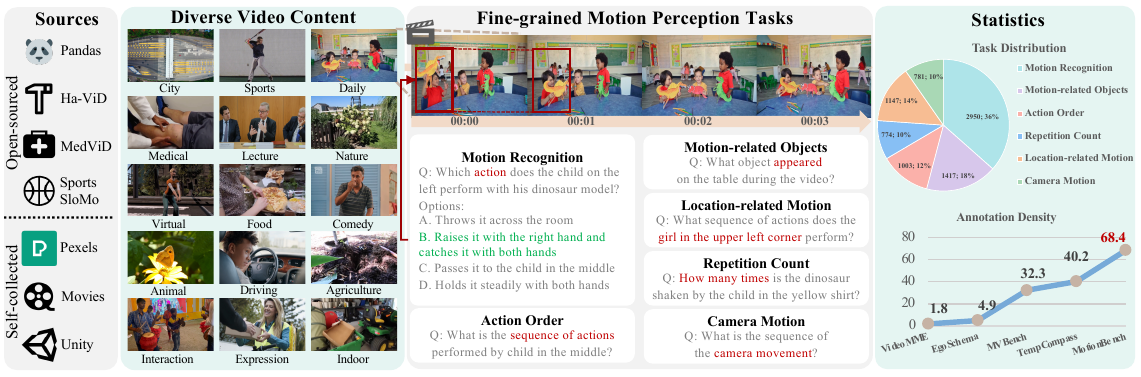}
    \caption{
    We propose MotionBench, a collection of manually curated multi-choice queries with video clips featuring dynamic changes from various scenes such as daily life and medical instructions. We devise six primary tasks to evaluate the capability of motion-level perception. Unlike previous story-level and event-level benchmarks, MotionBench is characterized by a significantly higher annotation density, allowing for the assessment of fine-grained motions.
    }
    \label{fig:pipe}
\end{figure*}

\begin{table*}[ht]
\small
\centering
\caption{The MotionBench curation process. Categories [1-3] refer to ``videos with intricate interactions", ``videos from specific fields" and ``virtual videos", detailed in \sref{subsec:data curation}. N. Vid/QA refers to the number of videos and queries in a category. min(\textrm{H}, \textrm{W}) is the minimum of the height and width of the video frames. \textit{len} refers to the processed video duration. We automatically construct the queries in Virtual scenes, and manually annotate the other QA pairs in MotinBench.}
\label{tab:data curation}
\begin{tabular}{cccccc}
\toprule
Category          & \# Videos/QAs    & Source         & Collection     & Post-process                                                  & Annotation     \\ \hline
\multirow{3}{*}{1} & \multirow{3}{*}{2,355/4,922} & Pexels         & Self-collected   & Directly adopt                                                & Caption \& QA        \\
                   & & Pandas-70M~\cite{chen2024panda}     & Open-sourced   & Segment with scene detection                                  & Caption \& QA          \\
                   & & Movie clips    & Self-collected & Segment with scene detection                                  & Caption \& QA         \\ \hline
\multirow{3}{*}{2} & \multirow{3}{*}{2,430/2,530}& MedVid~\cite{Gupta2022ADF}         & Open-sourced   & $\min(\textrm{H}, \textrm{W}) > 448$ \& len$\in[3, 60]$sec & QA                    \\
                   & & SportsSloMo~\cite{chen2024sportsslomo}            & Open-sourced   & $\min(\textrm{H}, \textrm{W})>448$ \& len$\in[3, 60]$sec & QA                     \\
                   & & HA-ViD~\cite{zheng2023havid}         & Open-sourced   & $\min(\textrm{H}, \textrm{W})>448$ \& len$\in[3, 60]$sec & QA                  \\ \hline
3                 & 600/600  & Virtual scenes & Self-collected & Remove renderings with occlusion                              & Automatic QA          \\ \bottomrule

\end{tabular}
\vspace{-1em}
\end{table*}
We introduce MotionBench, an evaluation benchmark designed to assess the motion-level perception capability of video VLMs. Fine-grained motion understanding is of paramount importance across a variety of daily scenarios, including human interaction, expression recognition, medical instruction, ambient object motion, sports replay, virtual reality, \etc Our approach begins with the collection of video clips from these diverse cases, which are then filtered and processed into the desired formats. We devise six primary categories of question types to evaluate the candidates' motion-level understanding, and we manually annotate the questions and answers within these categories, yielding the proposed MotionBench. \Tref{tab:data curation} provides an overview of our data construction pipeline.

\begin{figure*}[h]
    \centering
    \begin{subfigure}{0.2\textwidth}
            \centering
        \includegraphics[width=\textwidth]{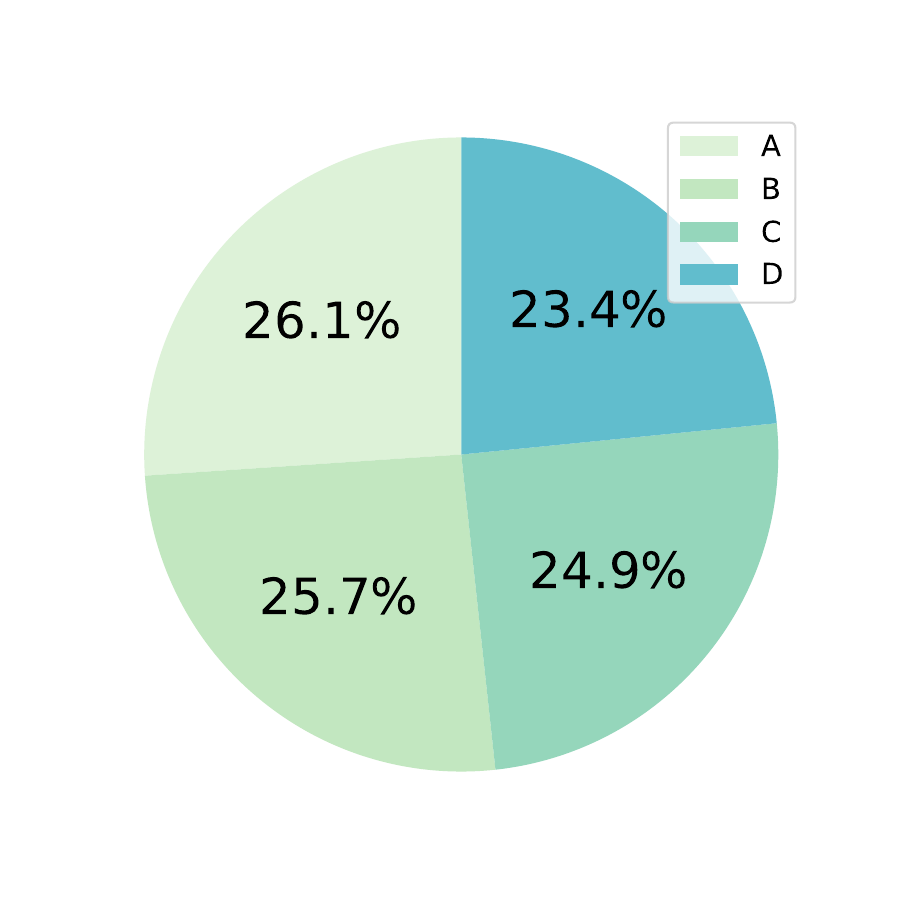}
        \caption{Option distribution}
        \label{fig:fig1}
        
    \end{subfigure}
    \begin{subfigure}{0.2\textwidth}
        \centering
        \includegraphics[width=\textwidth]{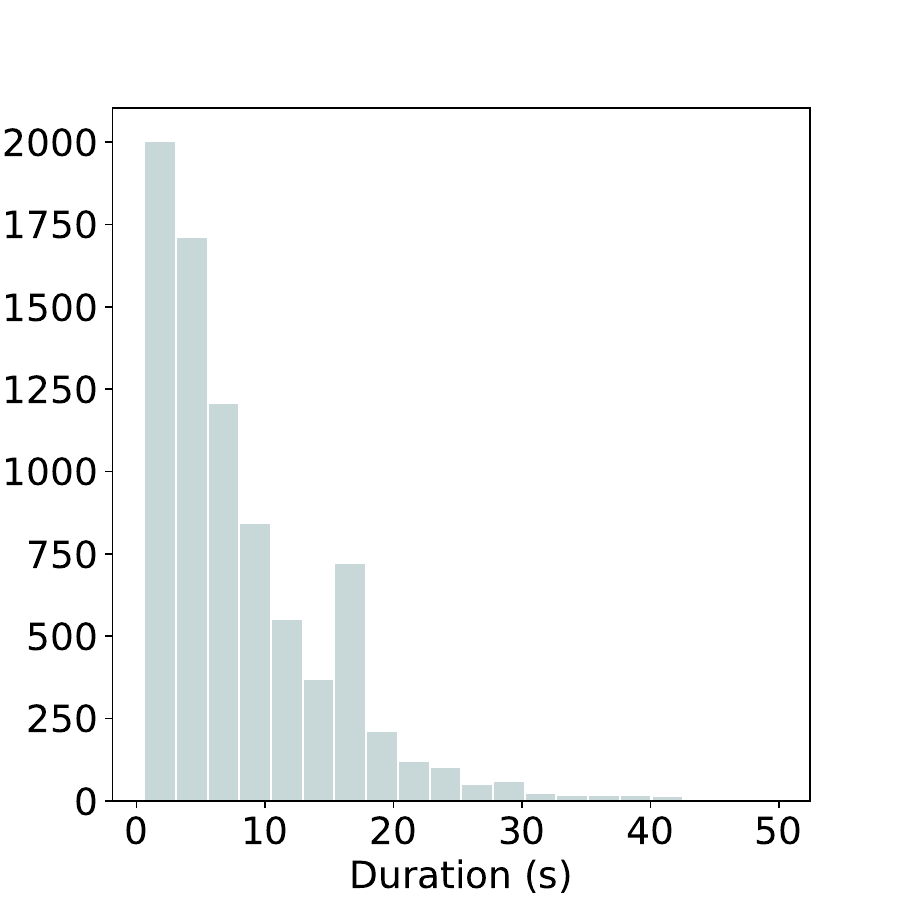}
        \caption{Video duration}
        \label{fig:fig3}
    \end{subfigure}
    \begin{subfigure}{0.2\textwidth}
        \centering
        \includegraphics[width=\textwidth]{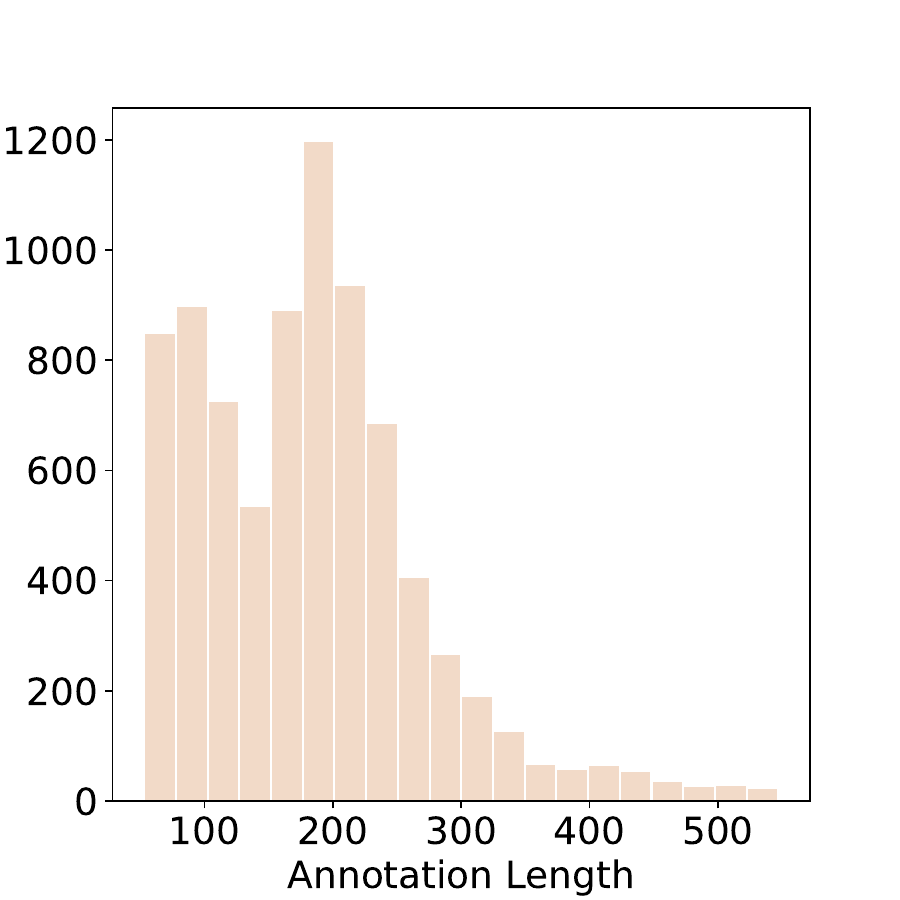}
        \caption{Annotation length}
        \label{fig:fig4}
    \end{subfigure}
    \begin{subfigure}{0.2\textwidth}
        \centering
        \includegraphics[width=\textwidth]{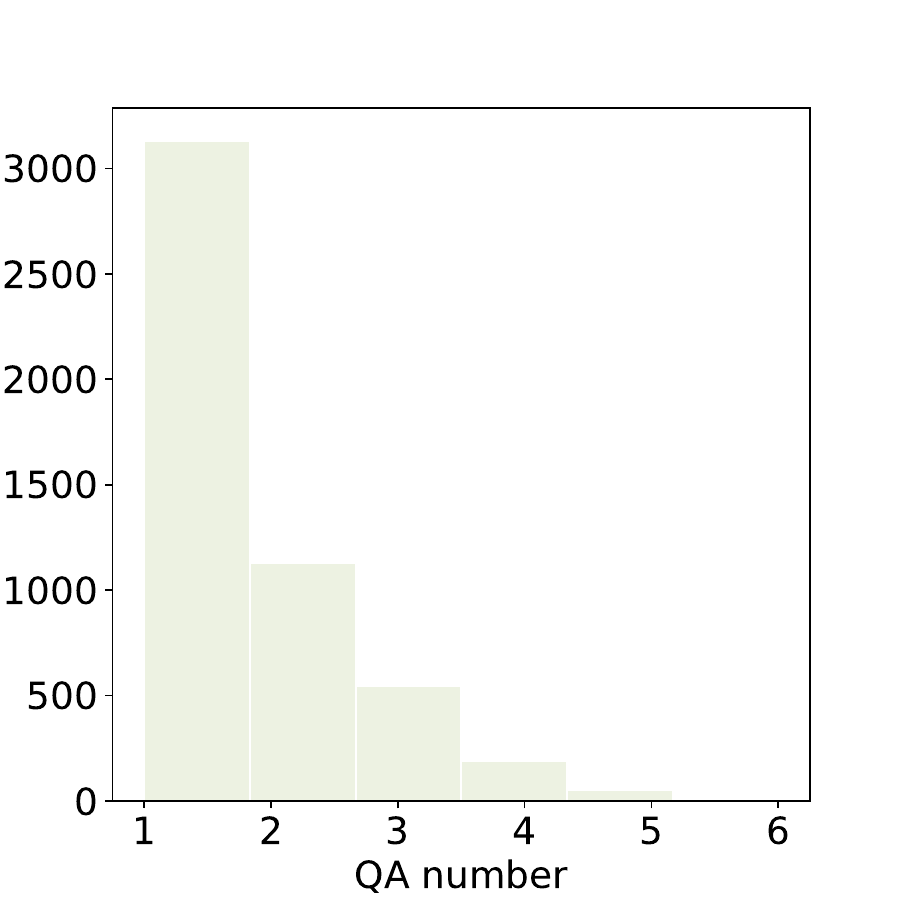}
        \caption{QA per video}
        \label{fig:fig2}
    \end{subfigure}
    \caption{Basic statistics of MotionBench. }
    \label{fig: base stat}
    \vspace{-1em}
\end{figure*}

\begin{figure}[t]
    \centering
    \includegraphics[width=0.46\textwidth]{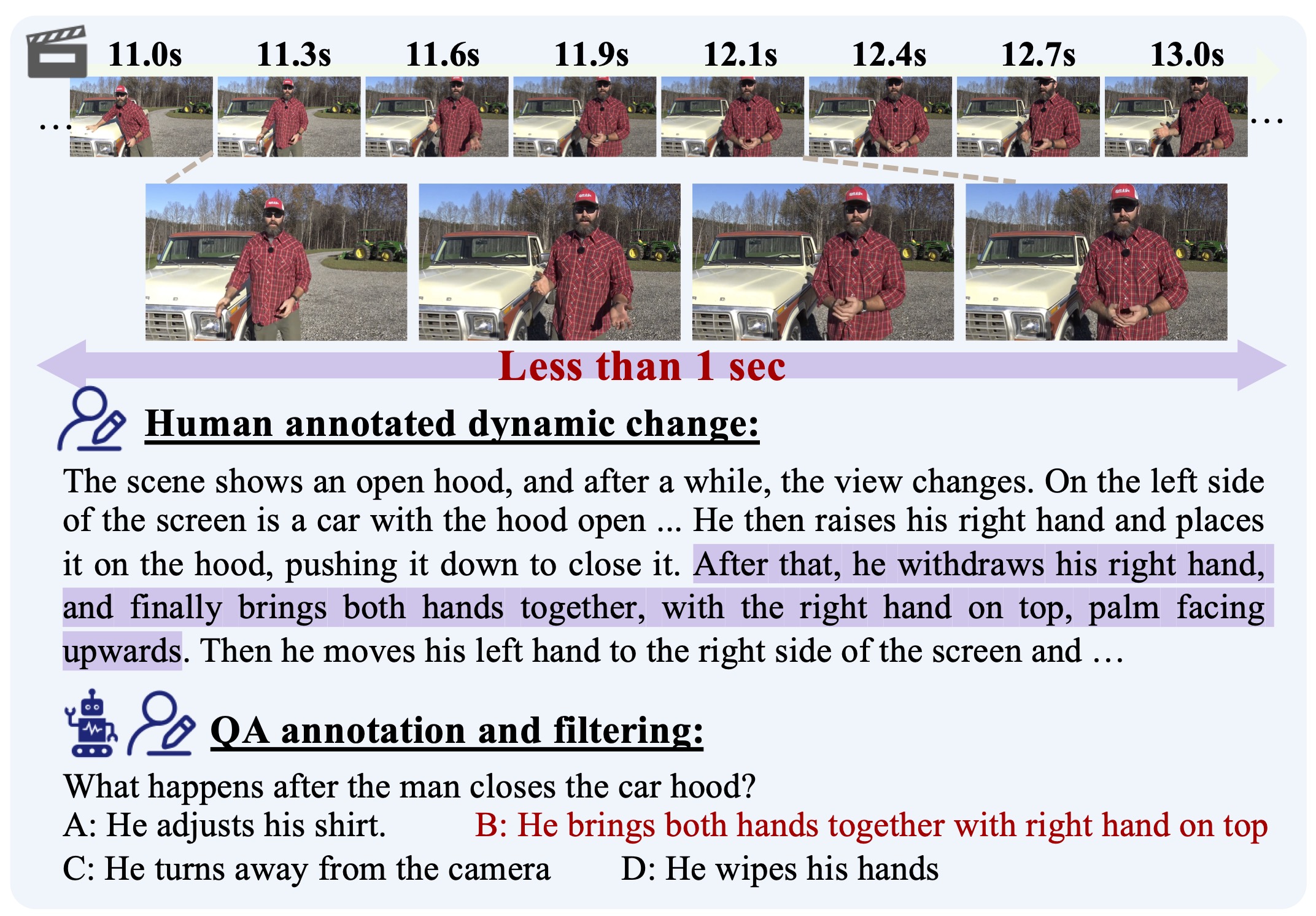}
    \caption{Example of dynamic information annotation}
    \label{fig:dy info anno}
    \vspace{-1em}
\end{figure}

\subsection{Data Curation}
\label{subsec:data curation}
In this section, we elaborate on the video curation, filtering, and annotation process.

\noindent \textbf{Video Collection.} 
We obtain raw videos from publicly available datasets as well as from our self-collected corpus. Based on the video sources, the vividness of the scenes, and the complexity of the scenarios, we split the videos into three distinct categories. Each category is processed and annotated using tailored pipelines accordingly:
\begin{itemize}
    \item \textbf{Videos with intricate interactions}: We acquire publicly-available videos from Panda-70M~\cite{chen2024panda} and Pexels\footnote{https://www.pexels.com} and collect high-quality movie clips featuring various actions and motions, attributing to a total of 2355 videos. To ensure uniformity in clip duration, we follow the methodology in Panda-70M~\cite{chen2024panda} to utilize a scene detection tool\footnote{https://github.com/Breakthrough/PySceneDetect} to segment these videos into event-level clips. 
    \item \textbf{Videos from specific fields}: We collect videos from MedVid~\cite{Gupta2022ADF}, SportsSloMo~\cite{chen2024sportsslomo} and Ha-ViD~\cite{zheng2023havid}, representing specific use cases in medical, sports and industrial applications. These videos usually consist of one or two simple motions and demonstrate less complicated interactions. For this category, we filter out videos longer than 60 seconds or resolutions less than $448\times448$ pixels. An amount of 2430 videos are retrieved in this category.
    \item \textbf{Synthetic videos}: The above-mentioned videos are mostly from real-world scenes. For further evaluation in virtual reality applications, we render avatars with simple motions using the Unity rendering engine. Furthermore, graphic engines generate renderings that exclusively focus on motion changes, making them highly suitable for the assessment of motion perception. We randomly sample 20 motions from a publicly available website\footnote{https://www.mixamo.com}, and select 6 avatars and 5 scenes to render virtual avatars from a pool of 15 different viewpoints. Renderings with occlusion are manually filtered. Please refer to the supplementary for details in rendering.
\end{itemize}

\noindent \textbf{Task Definition.} To assess the capability in motion-level perception, we propose six categories of questions. Examples and the distribution of each category are illustrated in \fref{fig:pipe}. A detailed description of each category is listed:
\begin{itemize}
    \item \textbf{Motion Recognition (MR)}: Questions focus on what kind of motion emerged in the given video clips.
    \item \textbf{Location-related Motion (LM)}: Questions assessing the relative location changes before and after the motion takes place, and queries regarding a specific location.
    \item \textbf{Action Order (AO)}: Complex actions are composed of a sequence of motions. Questions in this category focus on the order of these motions.
    \item \textbf{Repetition Count (RC)}: Certain subtle motions occur rapidly but are repeated multiple times, such as nodding or jumping. This category of questions evaluates the model's ability to recognize and interpret such motions.
    \item \textbf{Motion-related Objects (MO)}: Queries designed to evaluate the model's ability to identify small objects involved in motion interactions.
    \item \textbf{Camera Motion (CM)}: Questions focus on the camera motion changes and trajectory, including the order and combinations of different motion types.
\end{itemize}

\noindent \textbf{Question Answer Annotation.}
We employ different annotation pipelines for the above-mentioned video categories. For videos with intricate interactions, it is impractical to directly annotate the whole video clip, since the total complexity and quantity of the motions are too large. Therefore, we first manually annotate these videos with captions that focus on the dynamic changes within the video. Subsequently, we prompt GPT-4o~\cite{gpt4} to generate 6 question-answer pairs for each video clip. For the prompt template and more details, please refer to the supplementary material. We find that the generated QA pairs are not only diverse in type but also presented in various sentence structures. We show an example of the dynamic information annotation pipeline in \fref{fig:dy info anno}.

In addition, we also drop all the questions that can be answered solely based on common knowledge and a single frame. We use various \textbf{image} VLMs to predict answers using the first frame as input and discard questions that are answered correctly by all VLMs. Then, we manually filter out any questions with incorrect phrasing or ambiguous answers and categorize them. Finally, 4922 queries and answers are retained. 

For videos from specific fields, we directly annotate the questions within the designed task types. A total of 2530 QA pairs are selected. 
For virtual videos, where we already possess the ground truth annotations for each query, we automatically construct the questions and corresponding options. Finally, 600 QA pairs are generated.

\noindent \textbf{Evaluation protocal.} we use regular expression matching to identify the first uppercase letter in the response as the prediction. We append ``Only reply with the best option." to the end of each question for instruction following.

\subsection{Dataset Statistics}

 \begin{figure*}[t]
    \centering
    \includegraphics[width=\linewidth]{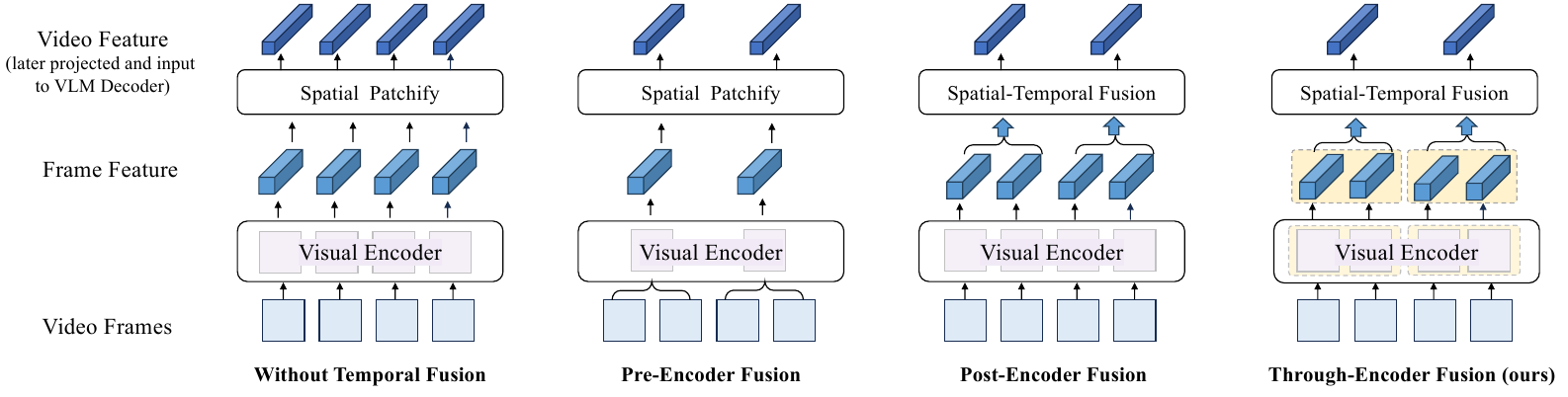}
    \caption{Summarization of prevalent paradigms for video compression and our proposed Through-Encoder Fusion (TE Fusion). Here we only illustrate the part before the VLM decoder where temporal compression performs.}
    \vspace{-1em}
    \label{fig:model_comparison}
\end{figure*}

MotionBench consists of 5385 videos and 8052 QAs, and each QA pair consists of a question, four options, an answer, and a category. The task distribution is displayed in \fref{fig:pipe}.

\noindent \textbf{Annotation Density.}
MotionBench is designed especially for evaluating the video VLM's motion-level perception capability. Such evaluation requires a larger annotation density per second. We define ``Annotation Density" to represent such attribute, defined as follows: 
\begin{equation}
    \mathrm{Annotation\, Density} = \frac{\mathrm{Total\,length\,of\,questions}}{\mathrm{Video\,duration}}
\end{equation}
The results are demonstrated in \fref{fig:pipe}. MotionBench features an Annotation Density of 68.4, which is two times more than existing benchmarks. 

\noindent \textbf{Basic Statistics.}
In \fref{fig: base stat}, we illustrate the distribution of options, number of QAs per video, duration, and annotation length in the MotionBench. Regarding the distribution of answer options in MotionBench, it can be observed that the various options generally adhere to a random distribution. Due to our manual removal of erroneous and overly simplistic questions, it can be seen that the QA pairs in ``Videos with intricate interactions" have been thoroughly filtered, resulting in the elimination of nearly half of the QA data. The video lengths in MotionBench are primarily concentrated around under 10 seconds, as motion events usually occur in very brief segments of the videos.

\section{Model Design: Motion-Level Perception}

Motion-level video perception demands high-frame-rate input, while the maximum input frame rate is significantly constrained by the sequence length limitations of VLMs, which are bounded by both infrastructure and computational budgets during training and inference. Therefore, it's necessary to design an efficient video understanding model structure with dense video representation. 
Recent studies, particularly in the domain of long video understanding, introduce various types of video feature compression methods~\citep{liu2024kangaroo, xu2024pllava, wang2024qwen2, wang2024internvideo2}, but lack comprehensive and fair comparisons across all methods. Therefore, We comprehensively investigate commonly used architectures for video compression and categorize prevalent paradigms in \cref{fig:model_comparison}.

\begin{itemize}
    \item \textbf{Without Temporal Fusion}: A baseline widely used in \citep{hong2024cogvlm2, zhang2024llavanextvideo}. Each frame is independently processed by the visual encoder and projected into the decoder space. 
    \item \textbf{Pre-Encoder Fusion}: 
    This architecture conducts temporal fusion among neighboring $k$ frames before the visual encoder, usually in pixel space.  
    The temporal fusion operator varies across implementations. Typical examples include Qwen2-VL~\citep{wang2024qwen2} where two adjacent frames are concatenated along the channel dimension for joint processing, and \citet{kim2024image} which merges several nearby frames into a single image.
    \item \textbf{Post-Encoder Fusion}: In this architecture, each frame first \textit{independently} goes through the visual encoder to generate frame-specific features, then performs feature fusion among neighboring frames with spatial-temporal fusion modules. Note that no temporal relationships are captured during visual encoding. This paradigm is the most widely adopted in video architecture with compression, with multiple variations in temporal fusion operators such as adaptive pooling~\citep{xu2024pllava}, QFormer~\citep{li2023blip}~\citep{wang2024internvideo2}, and unified spatial-temporal patchification~\citep{liu2024kangaroo}. 
\end{itemize}

\begin{table*}[t]
\centering
\small
\caption{Evaluation results of the existing video VLMs. Abbreviations: MR (Motion Recognition), LM (Location-related Motion), CM (Camera Motion), MO (Motion-related Objects), AO (Action Order), RC (Repetition Count). We randomly split MotionBench into ``dev" and ``test". We will release the ground truth answers in the ``dev" set and set up an online platform for results submission in the ``test" set.}
\resizebox{\textwidth}{!}{%
\setlength{\tabcolsep}{3mm}{
\label{tab:dataset eval}
\begin{tabular}{ccccccccccc}
\hline
Model          & LLM                  & \multicolumn{1}{c|}{\# Frames} & \multicolumn{1}{c|}{\begin{tabular}[c]{@{}c@{}}Dev AVG \\ (4020)\end{tabular}} & \begin{tabular}[c]{@{}c@{}}Test AVG \\ (4034)\end{tabular} & MR   & LM   & CM   & MO   & AO   & RC   \\ \hline
\color{dt}Random         & \color{dt}-        & \multicolumn{1}{c|}{\color{dt}-}    & \multicolumn{1}{c|}{\color{dt}0.25}         & \color{dt}0.25         & \color{dt}0.25 & \color{dt}0.25 & \color{dt}0.25 & \color{dt}0.25 & \color{dt}0.25 & \color{dt}0.25 \\ \hline
\multicolumn{11}{c}{\it{LLM: \textbf{Text} as Input}}                                                                                                                                                                                             \\ \hline
GPT-4o~\cite{gpt4}         & -               & \multicolumn{1}{c|}{-}         & \multicolumn{1}{c|}{0.33}                                                      & 0.33                                                       & 0.31 & 0.34 & 0.36 & 0.37 & 0.42 & 0.23 \\ \hline
\multicolumn{11}{c}{\it{Video VLMs : \textbf{Text + Multiple Frames} as Input}}                                                                                                                                                                                              \\ \hline
Gemini 1.5 Pro~\cite{reid2024gemini} & -               & \multicolumn{1}{c|}{1fps}      & \multicolumn{1}{c|}{0.51}                                                      & 0.50                                                        & 0.51 & 0.52 & 0.54 & 0.67 & 0.40 & 0.22 \\
Qwen2VL-2B~\cite{Qwen2VL}      & Qwen2~\cite{wang2024qwen2} & \multicolumn{1}{c|}{1fps}      & \multicolumn{1}{c|}{0.48}                                                      & 0.47                                                       & 0.49 & 0.49 & 0.42 & 0.62 & 0.32 & 0.28 \\
Qwen2VL-7B~\cite{Qwen2VL}      & Qwen2~\cite{wang2024qwen2} & \multicolumn{1}{c|}{1fps}      & \multicolumn{1}{c|}{0.52}                                                      & 0.52                                                       & 0.52 & 0.55 & 0.49 & 0.68 & 0.39 & 0.32 \\
Qwen2VL-72B~\cite{Qwen2VL}     & Qwen2~\cite{wang2024qwen2} & \multicolumn{1}{c|}{1fps}      & \multicolumn{1}{c|}{0.57}                                                      & \textbf{0.58}                                                       & 0.58 & \textbf{0.61} & \textbf{0.63} & 0.72 & \textbf{0.47} & 0.31 \\
InternVL-40B~\cite{internvl2} & NH-2-Yi-34B~\cite{NH2Yi34B} & \multicolumn{1}{c|}{8}         & \multicolumn{1}{c|}{0.55}                                                      & 0.54                                                       & 0.54 & 0.58 & 0.49 & \textbf{0.76} & 0.41 & 0.30  \\
PLLaVA-34B~\cite{xu2024pllava} & Yi-34B~\cite{NH2Yi34B}    & \multicolumn{1}{c|}{16}        & \multicolumn{1}{c|}{0.52}                                                      & 0.51                                                       & 0.55 & 0.51 & 0.47 & 0.66 & 0.38 & 0.31 \\
CogVLM2-Video~\cite{hong2024cogvlm2}  & LLaMA3-8B~\cite{llama3modelcard}  & \multicolumn{1}{c|}{24}        & \multicolumn{1}{c|}{0.41}                                                      & 0.44                                                       & 0.43 & 0.39 & 0.38 & 0.64 & 0.37 & 0.33 \\
GLM-4V-plus~\cite{hong2024cogvlm2} & GLM4~\cite{glm2024chatglm}                 & \multicolumn{1}{c|}{30}        & \multicolumn{1}{c|}{0.54}                                                      & 0.55                                                       & 0.57 & 0.57 & 0.54 & 0.69 & 0.40  & 0.37 \\
LLaVA-NeXT~\cite{zhang2024llavanextvideo} & Yi-34B~\cite{NH2Yi34B}   & \multicolumn{1}{c|}{32}        & \multicolumn{1}{c|}{0.48}                                                      & 0.40                                                        & 0.53 & 0.45 & 0.36 & 0.66 & 0.39 & 0.23  \\
MiniCPM-V2.6~\cite{yao2024minicpm}   & Qwen2~\cite{wang2024qwen2}    & \multicolumn{1}{c|}{64}        & \multicolumn{1}{c|}{0.52}                                                      & 0.53                                                       & 0.56 & 0.49 & 0.45 & 0.72 & 0.39 & 0.33 \\
Oryx-34B~\cite{liu2024oryx}       & Yi-34B~\cite{NH2Yi34B} & \multicolumn{1}{c|}{64}        & \multicolumn{1}{c|}{0.49}                                                      & 0.49                                                       & 0.48 & 0.52 & 0.44 & 0.65 & 0.42 & 0.32 \\
TE Fusion (ours) & GLM4-9B~\cite{glm2024chatglm}              & \multicolumn{1}{c|}{16} & \multicolumn{1}{c|}{\textbf{0.58}}                                                      & \textbf{0.58}                                                       & \textbf{0.64} & 0.59 & 0.51 & 0.69 & 0.41 & \textbf{0.39} \\ \bottomrule
\end{tabular}
}}
\vspace{-1em}
\end{table*}

All compression architectures rely on the assumption that redundancy exists between frames which contributes little to the video’s comprehension and can therefore be removed. Achieving a higher compression ratio requires a more precise and thorough capture of this redundant information. 
However, current video temporal compression methods have a common limitation: the inter-frame relationships are considered only within the small compression operator, and each frame is treated independently before the operator. Consequently, it is difficult for this kind of shallow fusion to effectively capture higher-level redundancies. 
For instance, in a video of a running person, the individual’s position, posture, and even the camera angle vary continuously. 
Only by applying sophisticated inter-frame fusion techniques can the model unify their representation throughout the video and capture this higher-level redundancy. 
Based on this observation, we propose a novel Through-Encoder Fusion paradigm that introduces deeper fusion across neighboring frames:
\begin{itemize}
    \item \textbf{Through-Encoder Fusion (TE Fusion)}: During the visual encoding stage, adjacent frames are grouped in sets of $k$ and apply group-level self-attention. This design gives the capacity to compute temporal dependencies through the whole visual encoder and conduct deep fusion. Following this, spatial-temporal compression is performed on each group of $k$ frames.
\end{itemize}

Note that Through-Encoder Fusion represents a class of temporal compression methods that perform deep frame fusion before applying the compression operator. In this work, we experiment with the straightforward approach, leaving other variations for future exploration.

\section{Experiments}
\subsection{Evaluation on MotionBench}\label{sec5.1-eval-motionbench}

We comprehensively evaluate the performance of existing video VLMs' capability in motion-level perception on MotionBench. We include multiple models with various model sizes and VLMs. The results are listed in \Tref{tab:dataset eval}. TE Fusion represents our proposed model, which uses TE Fusion on GLM-4V-9B backbone, with 16 input frames and a compress ratio of 4. Among existing VLMs, Qwen2VL-72B achieves the best overall performance on the dev and test set and scores highest in 3 out of 6 categories. Surprisingly, TE Fusion achieves state-of-the-art results with a 9B LLM backbone, verifying the effectiveness of our method.

\noindent\textbf{Analysis. } With text input alone, GPT-4 achieves an accuracy rate of 0.3 to 0.4, surpassing the random baseline of 0.25. This result indicates that LLMs possess a prior probability for certain actions, even when based only on text (note that questions answerable purely by common knowledge are filtered out during data curation). Building on LLMs, video VLMs improve accuracy by just 0.05 to 0.2, highlighting that current video VLMs still face challenges in reliably recognizing even short, simple motions.
For the Repetition Count category, all models, except GLM-4V-9B with TE Fusion and GLM-4V-plus, scored near random. This is likely because fast motions are challenging to count at low frame rates or are easily overlooked by the models. Conversely, models generally achieved high scores in the Motion-related Objects category. This could be attributed to the pretraining video data, which is often constructed from image descriptions and emphasizes the objects in the video.

We further analyze the questions that all models fail to answer. The largest proportion involves fine-grained motion, suggesting that certain actions and their associated captions may be underrepresented in the training data. When examining questions by video duration, we find that even for short videos (0-4 sec), the proportion of all-model-failed questions remains 11\% to 14\%, highlighting models' difficulty in distinguishing certain motions even with limited content. As video duration increases, the failure rate rises significantly, reaching 18\% for videos longer than 18 seconds. 
Further analysis from more perspectives and case studies are provided in the appendix.

\begin{figure*}[ht]
    \centering
    \includegraphics[width=0.95\linewidth]{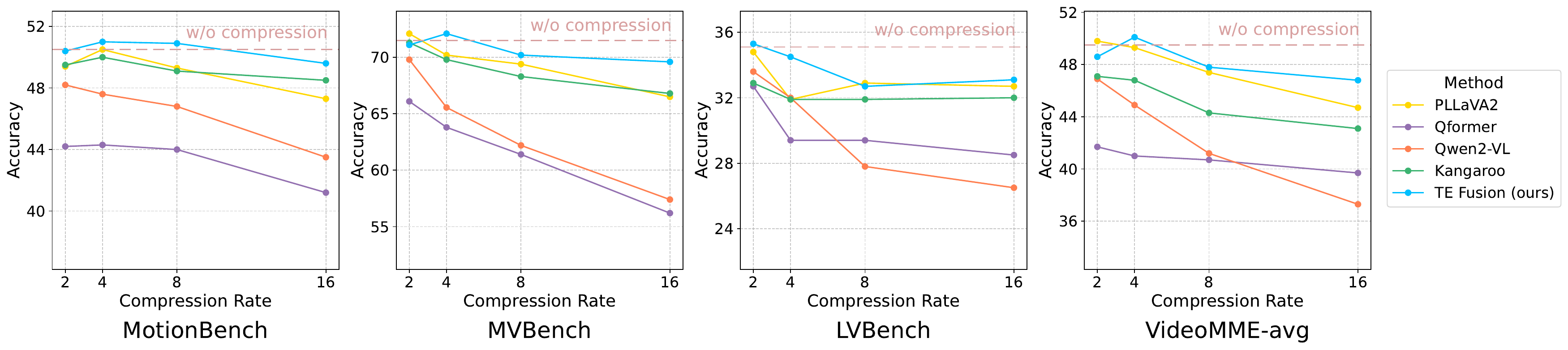}
    \caption{Model performance variation with respect to different compression ratios \( k = 2, 4, 8, 16 \), given a fixed VLM input frame count of \( N_\text{input} = 16 \). The pink dotted line represents the performance of the baseline model, which processes 16 frames without temporal compression. Note that each compression method is re-implemented on the GLM-4V-9B backbone to ensure a fair comparison.}
    \vspace{-1em}
    \label{fig:model_exp_all_plots}
\end{figure*}

\subsection{Experiments on Video Feature Compression} \label{sec5-2: comparison of video compression}

To comprehensively and fairly evaluate all paradigms of video compression architecture, we implement representative methods from each paradigm based on the same image foundation model, GLM-4V-9B~\citep{hong2024cogvlm2}: (1) Pre-encoder fusion: Qwen2-VL~\citep{wang2024qwen2}; (2) Post-encoder fusion: QFormer~\citep{li2023blip}, PLLaVA~\citep{xu2024pllava}, Kangaroo~\citep{liu2024kangaroo}; (3) Through-encoder fusion: our proposed implementation; (4) Baseline without temporal fusion. All models take $224\times224$-pixel input and are trained for 10,000 iterations with a global batch size of 768 on the same collection of open-source datasets. Note that the training data is a subset of the data used in \cref{sec5.1-eval-motionbench}. The details of training and architecture are further provided in the Appendix. Besides \benchname(dev), our motion-level video benchmark, we further evaluate all models on MVBench~\citep{li2024mvbench}, LVBench~\citep{wang2024lvbench}, and Video-MME~\citep{fu2024video} as the representation of video benchmarks of varying duration and content.

Let \( N_{\text{input}} \) represent the number of frames fed into the visual encoder, and let each frame's uncompressed length at the VLM decoder be \( l \) tokens. With a given compression ratio \( k \), the total compressed input length for the VLM decoder is \( L_{\text{decoder}} = \frac{N_{\text{input}} \times l}{k} \). Our experiment centers on addressing two primary questions:
\begin{enumerate}
    \item For a fixed sequence length at the VLM decoder (\( L_{\text{decoder}} \)), how does performance vary as the compression ratio increases?
    \item For a fixed number of input frames (\( N_{\text{input}} \)), how does performance respond to changes in the compression ratio, and is there an optimal compression ratio?
\end{enumerate}

\begin{table}[t]
\caption{Benchmark results for different compression methods at various compression rates, all using the same sequence length in the VLM decoder. We set \(\frac{N_{\text{input}}}{k} = 4\), with the baseline representing video models that process 4 frames without compression. Note that each compression method is re-implemented on the GLM-4V-9B backbone to ensure a fair comparison.} \label{tab:compress_normal_main}
\vspace{-0.7em}
\resizebox{\columnwidth}{!}{%
\setlength{\tabcolsep}{1mm}{
\begin{tabular}{ccccccc}
\toprule
\multirow{2}{*}{\begin{tabular}[c]{@{}c@{}}$k$\end{tabular}} & \multirow{2}{*}{Method} & \multirow{2}{*}{\small{MotionBench}} & \multirow{2}{*}{\small{MVBench}} & \multicolumn{3}{c}{\small{VideoMME}}                  \\ \cline{5-7} 
                                                                         &                         &                              &                          & \small short         & \small medium        & \small long          \\ \hline
1                                                                        & \small baseline                & 47.6                         & 64.5                     & 51.4          & 41.0          & 38.3          \\ \hline
\multirow{5}{*}{2}                                                       & \small QFormer                 & 43.5                         & 62.1                     & 42.8          & 39.6          & 36.3          \\
                                                                         & \small Qwen2-VL                & 48.0                         & 66.5                     & 54.1          & 43.1          & 37.8          \\
                                                                         & \small PLLaVA                  & 48.5                        & 68.8                     & 54.9          & 44.9          & 39.6          \\
                                                                         & \small Kangaroo                & 48.4                         & \textbf{69.2}            & 55.4          & 43.0          & 38.8          \\
                                                                         & \small TE Fusion (ours)        & \textbf{49.1}                         & 69.0                     & \textbf{55.2} & \textbf{46.3} & \textbf{40.0} \\ \hline
\multirow{5}{*}{4}                                                       & \small QFormer                 & 44.3                         & 63.8                     & 45.2          & 41.0          & 36.8          \\
                                                                         & \small Qwen2-VL                & 47.6                         & 65.6                     & 51.8          & 43.4          & 39.4          \\
                                                                         & \small PLLaVA                  & 50.5                         & 70.2                     & 58.9          & 46.4          & 41.3          \\
                                                                         & \small Kangaroo                & 50.0                         & 69.8                     & 55.3          & 45.6          & 39.5          \\
                                                                         & \small TE Fusion (ours)        & \textbf{51.0}                         & \textbf{72.1}            & \textbf{61.0} & \textbf{47.3} & \textbf{42.1} \\ \bottomrule
\end{tabular}
}}
\vspace{-1.2em}
\end{table}

For the first question, we conduct experiments with \(\frac{N_{\text{input}}}{k} = 4\) and \(8\), varying the compression rate \(k\) at \(2, 4, 6,\) and \(8\). Results for \(\frac{N_{\text{input}}}{k} = 4\) are shown in \cref{tab:compress_normal_main}, with complete results included in the Appendix due to space constraints. Given the same \(L_{\text{decoder}}\), most temporal compression methods demonstrate performance improvements across all benchmarks, with higher compression rates generally yielding better scores. Notably, PLLaVA, Kangaroo, and TE Fusion show relatively strong results, with our TE Fusion achieving the highest scores in 9 out of 10 metrics, improving upon the baseline by 11.8\% on MVBench and 18.7\% on VideoMME-short with \(k=4\). Qwen2-VL performs well with \(k=2\) but shows minimal improvement (or even a decline) with \(k=4\), likely due to the limited high-level compression capabilities of post-encoder fusion. QFormer, on the other hand, occasionally underperforms compared to the baseline, potentially due to the complexity of the additional module, which is challenging to optimize during the video compression training stage.

For the second question, we set the input frame count to \( N_{\text{input}} = 16 \) and test compression rates of \( k = 2, 4, 6, \) and \( 8 \) across all methods. The results, shown in \cref{fig:model_exp_all_plots} (with full numerical data in the appendix), reveal that while all methods experience some performance decline as the compression rate increases, our TE Fusion method exhibits almost no performance drop for \( k \leq 4 \). Even with a larger \( k = 16 \), the average performance reduction remains under 4\% compared to the high-consumption baseline without compression. 
Additionally, the performance decline caused by temporal compression is less significant in shorter-duration videos (MotionBench, MVBench) compared to longer-duration videos (LVBench), suggesting that high-frame-rate input offers greater potential for effective, high-ratio temporal compression. 
Interestingly, We find that TE fusion achieves the highest score with compression-4 instead of compression-2 in 3 of 4 datasets. An explanation is that a higher compression rate increases attention length within the ViT component while decreasing it in the LLM component. 
This finding suggests that the computational allocation in previous video VLMs may be suboptimal and enlightens a new direction to improve model performance.

\section{Conclusion}

We present MotionBench, a new benchmark for assessing fine-grained motion understanding in video models. Our experiments show that current state-of-the-art models struggle with motion-level comprehension, emphasizing the need for specialized benchmarks. To tackle this, we propose the Through-Encoder (TE) Fusion method, which improves video feature representation by deeply integrating fusion within the visual encoder. TE Fusion achieves state-of-the-art results, especially under high compression, paving the way for advances in motion perception.

\subsubsection*{Acknowledgments}
This work is supported by Natural Science Foundation of China (NSFC) 62425601 and 62495063, Daimler Greater China Ltd. and Tsinghua University Joint Institute for Sustainable Mobility and the New Cornerstone Science Foundation through the XPLORER PRIZE. We thank Xiaohan Zhang, Yuean Bi, Xiaoying Ling, Jiapeng Wang, Zikang Wang from Zhipu AI for managing the data annotation team, and Zhao Xue from Zhipu AI for data management.

{
    \small
    \bibliographystyle{ieeenat_fullname}
    \bibliography{main}
}

\clearpage
\maketitlesupplementary

\section{Training Details}
Here we provide the detailed training hyperparameters for both TE Fusion in \cref{tab:dataset eval} and all ablated models in \cref{tab:compress_normal_main} and \cref{fig:model_exp_all_plots}. 

\begin{small}
\begin{table}[htbp]
\centering
    \caption{
        Training settings TE Fusion in \cref{tab:dataset eval} and all ablated models in \cref{tab:compress_normal_main} and \cref{fig:model_exp_all_plots}. 
    }
    \renewcommand{\arraystretch}{1.15}

    \setlength{\tabcolsep}{15mm}{
    \resizebox{0.8\columnwidth}{!}{
    \setlength{\tabcolsep}{10pt}
    \begin{tabular}{cc}
        \toprule
        \textbf{}             & \textbf{Configurations} \\ \midrule
        Total steps           & 10,000                  \\
        Warmup steps          & 1,000                   \\
        Global batch size     & 768                     \\
        Learning rate         & 8e-6                    \\
        Minimal learning rate & 1e-6                    \\
        Learning rate decay   & cosine                  \\
        Optimizer             & Adam                    \\
        Adam $\epsilon$       & 1e-8                    \\
        Adam $\beta$1         & 0.9                     \\
        Adam $\beta$2         & 0.95                    \\
        Precision             & bf16                    \\ \bottomrule
        \end{tabular}    }
    }

    \vspace{2mm}
  \label{tab:appendix_hyperparam_pretrain}
\end{table}
\end{small}

The training is conducted on several datasets, mainly including VideoChat~\cite{videochat}, VideoChatGPT~\cite{maaz2023videovhatgpt}, NExT-QA~\cite{nextqa}, CLEVRER~\cite{clevrer}, Kinetics-710~\cite{uniformerv2}, SthSthV2~\cite{sth}, Ego4D~\cite{ego4d}, TGIF-QA~\cite{tgif_qa}, WebVidQA~\cite{just_ask}, In-house VideoQA Dataset. We also include an in-house video QA dataset for better temporal understanding. 

\section{Model Details}

\textbf{A detailed explanation of TE Fusion}: Consider the temporal/spatial downsampling factor $k_t$ / $k_s$. The input video 
is first patchified into $X_\text{video} \in \mathbb{R}^{T \times h \times w \times d}$. Every $k_t$ frames are grouped along the spatial dimension, forming $X_\text{in} \in \mathbb{R}^{\frac{T}{k_t} \times (k_t hw) \times d}$.
$X_\text{in}$ is processed by a ViT-based visual encoder, where self-attention operates within each group of $k_t$ frames.
Note that the group-by-$k_t$ attention is only one implementation of deep fusion across $k_t$ frames, thus TE Fusion can be extended to a paradigm of architectures. 
The encoder outputs:  
{\small\[
X_\text{out} = \text{stack}(\text{enc}(X_\text{in}^{1:k_t}), \text{enc}(X_\text{in}^{k_t+1:2k_t}), ..., \text{enc}(X_\text{in}^{T-k_t+1:T})),
\]}
\noindent where $X_\text{in}^{(i \cdot k_t + 1):((i+1)\cdot k_t)} \in \mathbb{R}^{(k_t hw) \times d}$ represents the $i$-th group, and $X_\text{out} \in \mathbb{R}^{\frac{T}{k_t} \times (k_t  h  w) \times d} $.


Then, $X_{out}$ is reshaped into $\mathbb{R}^{T \times h \times w \times d}$ and processed by a 3D convolutional operator with kernel size ($k_t$, $k_s$, $k_s$) and stride ($k_t$, $k_s$, $k_s$), producing downsampled features $X_\text{feat} \in \mathbb{R}^{\frac{T}{k_t} \times \frac{h}{k_s} \times \frac{w}{k_s} \times d_\text{LLM}}$, which are fed into the LLM. This grouped-by-$k_t$ attention and $k_t$-downsampling allow deeper fusion of adjacent $k_t$ frames, effectively capturing inter-frame dynamics and redundancy, therefore retaining critical features for subsequent processing.

\textbf{Comparison among different compression architectures}: Assume the temporal compression ratio be $K$, The specific feature of each ablated architecture is:
\begin{enumerate}
\item TE-Fusion (ours): Before the visual encoder, we concatenate every neighboring $K$ frames into one sequence, and conduct self-attention across each $K$ frames to fuse temporal feature. After the visual encoder, the tokens of K frames are concatenated along the hidden-size dimension, downsampled and projected to the output dimension. 
\item Qwen2-VL: The neighboring $K$ frames are concatenated along the channel dimension and patchified into one feature. Afterward, they go through the visual encoder as a whole. Since the fusion is conducted in the pixel space before any feature extraction or fusion, the optimized temporal compression ratio is usually low, with a vast information loss if a large $K$. 
\item Kangaroo: This approach is the most similar one to TE Fusion, except that every frame is computed independently within the visual encoder and concatenated along the hidden size dimension to perform temporal downsample (with an MLP layer).
\item QFormer: After going through the visual encoder, the video feature is passed through a QFormer (learned from scratch). Every $K$ frames' feature is combined into a sequence to fusion temporal information within the QFormer. From the experiment, we found that, though being light-weighted, the QFormer is hard to optimize and model temporal relationships during the video instruction-tuning stage, resulting in poor performance. 
\item PLLaVA: This approach is similar to Kangaroo. Instead of fusion with the MLP layer, PLLaVA adopts a simple adaptive pooling. To avoid possible information loss, we conduct the pooling operation after the spatial downsample module. 
\end{enumerate}

To maintain a fair comparison, all model architectures are ablated with the same backbone, GLM-4V, with its model configuration as follows:

\begin{table}[h]
\centering
\caption{The model configurations of all ablated architectures.}
\setlength{\tabcolsep}{4mm}{
\resizebox{0.8\columnwidth}{!}{
\begin{tabular}{cc}
\toprule
\multicolumn{2}{c}{\textit{\textbf{VLM decoder}}}                      \\ \hline
Layers                         & 40                  \\
Hidden size                    & 4096                \\
Attention heads                & 32                  \\
num query groups               & 2                   \\
FFN hidden size                & 13696               \\
Sequence len                   & 4096                \\
Position embedding             & RoPE                \\
Normalization                  & RMSNorm             \\ \midrule
\multicolumn{2}{c}{\textit{\textbf{visual encoder}}} \\ \midrule
Input resolution               & 224                 \\
Patch size                     & 14                  \\
Post spatial downsample        & 2 $\times$ 2        \\
Layers                         & 63                  \\
Hidden size                    & 1792                \\
Attention heads                & 16                  \\ \bottomrule
\end{tabular}
}}
\end{table}

\section{QA Construction Process for Videos with Intricate Interactions}
Here we illustrate the QA generation process corresponding to \cref{fig:dy info anno}.
\subsection{Step1: Video caption annotation}
For videos with intricate interactions, it is impractical to directly annotate the whole video clip, since the total complexity and quantity of the motions are too large. Therefore, we first manually annotate these videos with captions that focus on the dynamic changes within the video 
(illustrated in \cref{fig:dy info anno}). We hired 15 adult annotators with at least a bachelor's degree and conducted annotations over 20 working days. Each annotator's daily salary was approximately 250 RMB. All annotations underwent a secondary review.

\subsection{Step2: Automatic QA generation}
Then we use GPT-4o to generate 6 questions corresponding to each video description. The instruction to GPT-4o emphasizes diversity as well as accuracy, as shown below:

~\

You are a professional question designer specializing in dynamic video details. Instead of a video, you will receive a detailed description of the first frame and all dynamic details throughout the video. Based on this description, design single-choice questions that focus on \textbf{the dynamic information} as if you’re viewing the video directly, using the two-dimensional categorization system below (Content Dimension, Question Logic Dimension).

\subsubsection*{Question Design Guidelines}
\begin{enumerate}
    \item Each question should have 4 options.
    \item For each question, combine one dimension from the Content Dimension and one from the Question Logic Dimension. It may draw from multiple highly related content dimensions.
    \item Focus only on representative and prominent events or actions to keep options clear and unique without being overly detailed or tricky. \textbf{Select the most fitting dimension combination} for each video and avoid repeated combinations where possible.
    \item Given possible ambiguities in some descriptions, \textbf{ensure the answer is unique and clear} to avoid deductions.
    \begin{itemize}
        \item \textbf{Ambiguity Example 1: Temporal ambiguity.} If a description reads, ``On the left, a woman in a khaki suit faces right, nodding her head while speaking. In the middle, a group faces the camera, and a man in a white shirt pulls a chair leftward to sit,'' the description is ambiguous and does not clarify the sequence of the woman’s actions and the man’s actions, making sequence ambiguous.
        \item \textbf{Ambiguity Example 2: Content ambiguity.} If the description states, ``The worker holds a long, thin tool,'' avoid options like ``screwdriver,'' as the tool could be any slender object.
    \end{itemize}
    \item Choose only prominent events or actions, avoiding minor or indeterminate details. Ensure each answer is \textbf{unique and clear}.
    \begin{itemize}
        \item \textbf{Minor Example:} If ``slightly bent elbow'' isn’t mentioned, it does not necessarily mean it did not happen; if the video says ``the mouth moved slightly a few times,'' it cannot be determined the interval and number of these movements, nor can it be determined whether the nose moved. Therefore, try to avoid using such minor actions for question creation or option design.
        \item Avoid subjective options, like ``Which detail reflects focus on work?'' unless a behavior clearly reflects it. Similarly, avoid terms like ``skilled movement'' or ``rhythmic.''
        \item Avoid overly similar distractors, e.g., ``chin moving up and down'' vs. ``slight opening and closing.''
    \end{itemize}
    \item Pretend you’re viewing the video, avoiding terms like ``based on the description'' or expressions related to the description text, including questions, options, and explanations.
    \item Aim for at least 4 questions to focus beyond appearance.
    \item Keep questions to around six, focusing only on representative events or actions and ensuring options are clear, unique, and straightforward.
    \item Questions should focus on dynamic actions only. The ``first frame description'' is supplementary and should not guide question design.
    \item The video dynamic information description does not contain causal or other logical relationships, therefore, do not involve logical relationships in the title.
\end{enumerate}

\subsubsection*{Categorization System}
\textbf{Content Dimension}
Below is the \textbf{Content Dimension} in the video classification system:
\begin{enumerate}
    \item \textbf{Human Dynamics:}
    \begin{enumerate}[label*=\arabic*.]
        \item Detailed actions of individuals
        \item Interaction among multiple people
        \item Emotional states and their changes
        \item Position and its changes (Location, Angle, etc.)
    \end{enumerate}
    \item \textbf{Object Dynamics:}
    \begin{enumerate}[label*=\arabic*.]
        \item Movement trajectory
        \item State changes
    \end{enumerate}
    \item \textbf{Animal Dynamics:}
    \begin{enumerate}[label*=\arabic*.]
        \item Detailed actions
        \item Position and its changes (Location, Angle, etc.)
    \end{enumerate}
    \item \textbf{Camera Movement:}
    \begin{enumerate}[label*=\arabic*.]
        \item Camera movement
    \end{enumerate}
    \item \textbf{Appearance Characteristics:}
    \begin{enumerate}[label*=\arabic*.]
        \item individuals
        \item objects
        \item environment
    \end{enumerate}
\end{enumerate}

\textbf{Question Logic Dimension}
Below is the \textbf{Question Logic Dimension} in the video classification system:

\begin{enumerate}
    \item Whether a movement occurs
    \item Movement count
    \item Sequence between multiple movements
    \item Appearance description and judgment
\end{enumerate}

\subsubsection*{Response Format}
Return only a Python list, where each element is a dictionary representing a question. Ensure it can be parsed by \texttt{json.loads()} without returning anything outside the list.

\subsection{VLM Filtering}
To avoid over simple QAs that do not utilize motion comprehension capability, we use various image VLMs to predict answers using the first frame as input and discard questions that are answered correctly by all VLMs. The VLMs include GPT-4o, Qwen2-VL, and GLM-4V-plus.

\subsection{Manual Check}
To ensure the correctness of all benchmark QAs, we further hire annotators to check all QAs generated by GPT-4o manually. A total of 10 annotators are hired to conduct manual checks for 5 days. The key points of inspection include: the reasonableness of the question, the correctness of the category, the relevance of the question to the video, the accuracy of the options, and the uniqueness of the correct answer. Each annotator's daily salary was approximately 250 RMB. All annotations underwent a secondary review.

\begin{table*}[h]
\caption{Benchmark results for different compression methods at various compression rates, all using the same sequence length in the VLM decoder. We set \(\frac{N_{\text{input}}}{k} = 4, 8\), with the baseline representing video models that process 4 frames without compression. Note that each compression method is re-implemented on the GLM-4V-9B backbone to ensure a fair comparison.} \label{tab:appendix-ablation-normal}
\small
\centering
\begin{tabular}{ccccccccc}
\toprule
\multirow{2}{*}{\begin{tabular}[c]{@{}c@{}}Equivalent\\ Frames $\frac{N_{\text{input}}}{k}$\end{tabular}} & \multirow{2}{*}{\begin{tabular}[c]{@{}c@{}}Compress\\ Rate\end{tabular}} & \multirow{2}{*}{Method} & \multirow{2}{*}{\begin{tabular}[c]{@{}c@{}}MotionBench\\ (dev)\end{tabular}} & \multirow{2}{*}{MVBench} & \multirow{2}{*}{LVBench} & \multicolumn{3}{c}{VideoMME}                  \\ \cline{7-9} 
                                                                                 &                                                                          &                         &                                                                              &                          &                          & short         & medium        & long          \\ \hline
\multirow{11}{*}{4}                                                              & 1                                                                        & baseline                & 47.6                                                                         & 64.5                     & 30.9                     & 51.4          & 41.0          & 38.3          \\ \cline{2-9} 
                                                                                 & \multirow{5}{*}{2}                                                       & QFormer                 & 43.5                                                                         & 62.1                     & 31.0                     & 42.8          & 39.6          & 36.3          \\
                                                                                 &                                                                          & Qwen2-VL                & 48.0                                                                         & 66.5                     & 31.5                     & 54.1          & 43.1          & 37.8          \\
                                                                                 &                                                                          & PLLaVA                  & 48.5                                                                         & 68.8                     & \textbf{33.4}            & 54.9          & 44.9          & 39.6          \\
                                                                                 &                                                                          & Kangaroo                & 48.4                                                                         & \textbf{69.2}            & 31.6                     & 55.4          & 43.0          & 38.8          \\
                                                                                 &                                                                          & TE Fusion (ours)        & \textbf{49.1}                                                                & 69.0                     & 32.3                     & \textbf{55.2} & \textbf{46.3} & \textbf{40.0} \\ \cline{2-9} 
                                                                                 & \multirow{5}{*}{4}                                                       & QFormer                 & 44.3                                                                         & 63.8                     & 29.4                     & 45.2          & 41.0          & 36.8          \\
                                                                                 &                                                                          & Qwen2-VL                & 47.6                                                                         & 65.6                     & 32.0                     & 51.8          & 43.4          & 39.4          \\
                                                                                 &                                                                          & PLLaVA                  & 50.5                                                                         & 70.2                     & 34.3                     & 58.9          & 46.4          & 41.3          \\
                                                                                 &                                                                          & Kangaroo                & 50.0                                                                         & 69.8                     & 31.9                     & 55.3          & 45.6          & 39.5          \\
                                                                                 &                                                                          & TE Fusion (ours)        & \textbf{51.0}                                                                & \textbf{72.1}            & \textbf{34.5}            & \textbf{61.0} & \textbf{47.3} & \textbf{42.1} \\ \hline
\multirow{11}{*}{8}                                                              & 1                                                                        & baseline                & 48.9                                                                         & 70.5                     & 32.9                     & 56.4          & 44.2          & 39.7          \\ \cline{2-9} 
                                                                                 & \multirow{5}{*}{2}                                                       & QFormer                 & 44.2                                                                         & 66.1                     & 32.7                     & 48.0          & 39.8          & 37.2          \\
                                                                                 &                                                                          & Qwen2-VL                & 48.2                                                                         & 69.8                     & 33.6                     & 57.3          & 44.1          & 39.4          \\
                                                                                 &                                                                          & PLLaVA                  & 49.4                                                                         & \textbf{72.1}            & 34.8                     & \textbf{61.0} & 46.4          & 39.8          \\
                                                                                 &                                                                          & Kangaroo                & 49.5                                                                         & 71.3                     & 32.9                     & 58.3          & 45.2          & 37.7          \\
                                                                                 &                                                                          & TE Fusion (ours)        & \textbf{50.4}                                                                & 71.1                     & \textbf{35.3}            & 58.7          & \textbf{46.9} & \textbf{40.2} \\ \cline{2-9} 
                                                                                 & \multirow{5}{*}{4}                                                       & QFormer                 & 44.4                                                                         & 66.0                     & 31.6                     & 45.7          & 40.0          & 37.2          \\
                                                                                 &                                                                          & Qwen2-VL                & 48.7                                                                         & 69.3                     & 33.1                     & 55.2          & 43.3          & 38.1          \\
                                                                                 &                                                                          & PLLaVA                  & 49.4                                                                         & 71.5                     & \textbf{36.2}            & 60.3          & 47.3          & 41.1          \\
                                                                                 &                                                                          & Kangaroo                & 49.9                                                                         & \textbf{71.6}            & 33.5                     & 59.0          & 45.8          & 38.2          \\
                                                                                 &                                                                          & TE Fusion (ours)        & \textbf{50.5}                                                                & \textbf{71.6}            & 36.0                     & \textbf{63.0} & \textbf{47.9} & \textbf{41.5} \\ \bottomrule
\vspace{-1em}
\end{tabular}
\end{table*}

\begin{table*}[t]
\caption{Model performance variation with respect to different compression ratios \( k = 2, 4, 8, 16 \), given a fixed VLM input frame count of \( N_\text{input} = 16 \). Note that each compression method is re-implemented on the GLM-4V-9B backbone to ensure a fair comparison.} \label{tab:appendix-16frame}
\small
\centering
\begin{tabular}{cccccccc}
\toprule
\multirow{2}{*}{Method}                                                     & \multirow{2}{*}{Compress Rate} & \multirow{2}{*}{MotionBench} & \multirow{2}{*}{MVBench} & \multicolumn{3}{c}{Video-MME} & \multirow{2}{*}{LVBench} \\ \cline{5-7}
                                                                            &                                &                              &                          & short    & medium    & long   &                          \\ \hline
w/o compression                                                             & 1                              & 50.5                         & 71.5                     & 60.7     & 46.6      & 41.1   & 35.1                     \\ \hline
\multirow{4}{*}{PLLaVA}                                                     & 2                              & 49.4                         & 72.1                     & 61.0     & 46.4      & 42.0   & 34.8                     \\
                                                                            & 4                              & 50.5                         & 70.2                     & 58.9     & 47.6      & 41.3   & 31.9                     \\
                                                                            & 8                              & 49.3                         & 69.4                     & 56.7     & 45.2      & 40.4   & 32.9                     \\
                                                                            & 16                             & 47.3                         & 66.5                     & 52.4     & 42.8      & 39.0   & 32.7                     \\ \hline
\multirow{4}{*}{QFormer}                                                    & 2                              & 44.2                         & 66.1                     & 48.0     & 39.8      & 37.2   & 32.7                     \\
                                                                            & 4                              & 44.3                         & 63.8                     & 45.2     & 41.0      & 36.8   & 29.4                     \\
                                                                            & 8                              & 44.0                         & 61.4                     & 45.3     & 40.6      & 36.3   & 29.4                     \\
                                                                            & 16                             & 41.2                         & 56.2                     & 44.2     & 39.4      & 35.4   & 28.5                     \\ \hline
\multirow{4}{*}{Qwen2-VL}                                                   & 2                              & 48.2                         & 69.8                     & 57.3     & 44.1      & 39.4   & 33.6                     \\
                                                                            & 4                              & 47.6                         & 65.6                     & 51.8     & 43.4      & 39.4   & 32.0                     \\
                                                                            & 8                              & 46.8                         & 62.2                     & 47.2     & 39.9      & 36.4   & 27.8                     \\
                                                                            & 16                             & 43.5                         & 57.4                     & 38.9     & 37.6      & 35.3   & 26.5                     \\ \hline
\multirow{4}{*}{Kangaroo}                                                   & 2                              & 49.5                         & 71.3                     & 58.3     & 45.2      & 37.7   & 32.9                     \\
                                                                            & 4                              & 50.0                         & 69.8                     & 55.3     & 45.6      & 39.5   & 31.9                     \\
                                                                            & 8                              & 49.1                         & 68.3                     & 51.9     & 42.3      & 38.7   & 31.9                     \\
                                                                            & 16                             & 48.5                         & 66.8                     & 49.8     & 42.4      & 37.1   & 32.0                     \\ \hline
\multirow{4}{*}{\begin{tabular}[c]{@{}c@{}}TE Fusion\\ (ours)\end{tabular}} & 2                              & 50.4                         & 71.1                     & 58.7     & 46.9      & 40.2   & 35.3                     \\
                                                                            & 4                              & 51.0                         & 72.1                     & 61.0     & 47.3      & 42.1   & 34.5                     \\
                                                                            & 8                              & 50.9                         & 70.2                     & 56.6     & 45.8      & 41.1   & 32.7                     \\
                                                                            & 16                             & 49.6                         & 69.6                     & 54.8     & 45.8      & 39.8   & 33.1                     \\ \bottomrule
\end{tabular}

\end{table*}


\section{Copyrights}
MotionBench is a research preview intended for non-commercial use only. For existing open-sourced video sources~\cite{zheng2023havid, chen2024sportsslomo, gupta2023dataset, chen2024panda}, we have carefully signed their provided license and will not re-distribute their videos without permission. For videos from Pexels, we will mandatorily ask the users to sign an agreement that the videos in MotionBench can only be used in non-commercial research and cannot be re-distributed. For self-collected movie clips, we will not directly distribute the raw videos, and will alternatively provide the download links and processing scripts.

\section{The originality of MotionBench.} 
Unlike former works that primarily focus on fixed action labels, limited gestures, or predefined video/action fields (\eg, egocentric scenes, sports, MotionBench aims to highlight motion-level understanding \emph{across general domain with large-scale QAs}.
Key innovations include:
\begin{enumerate}
\item Various QAs are uniformly distributed across diverse video sources, descriptions and motion object(s) for all categories.
In contrast, MVBench evaluates limited video sources and scenes: subcategories FA, MD and OI each rely only on a single data source, and
MD features only simulated movements of simple objects; 
\item  We process and annotate the raw videos from scratch with high annotation length (Fig.3a), without leveraging any existing annotations. MVBench heavily relies on previous datasets' annotations, which offers little incremental information and poses the danger of training data leakage; 
\item A semi-automated pipeline for discovering and annotating motions from the video sources. MVBench uses a rule-based pipeline without human corrections.
\end{enumerate}

\section{More Experimental Results}
Given the same sequence length in the VLM decoder, we benchmark results for different compression methods at various compression rates. We conduct experiments with \(\frac{N_{\text{input}}}{k} = 4\) and \(8\), varying the compression rate \(k\) at \(2, 4, 6,\) and \(8\). 
\cref{tab:appendix-ablation-normal} provide the complete results. 

Given the same VLM input frame count, we experiment different compression ratios on various architectures, with the numerical results illustrated in \cref{tab:appendix-16frame}.

\section{Case Study on Model Performance}
We show more case studies regarding the performance of existing models on MotionBench.

\begin{figure}[t]
    \centering
    \includegraphics[width=0.99\linewidth]{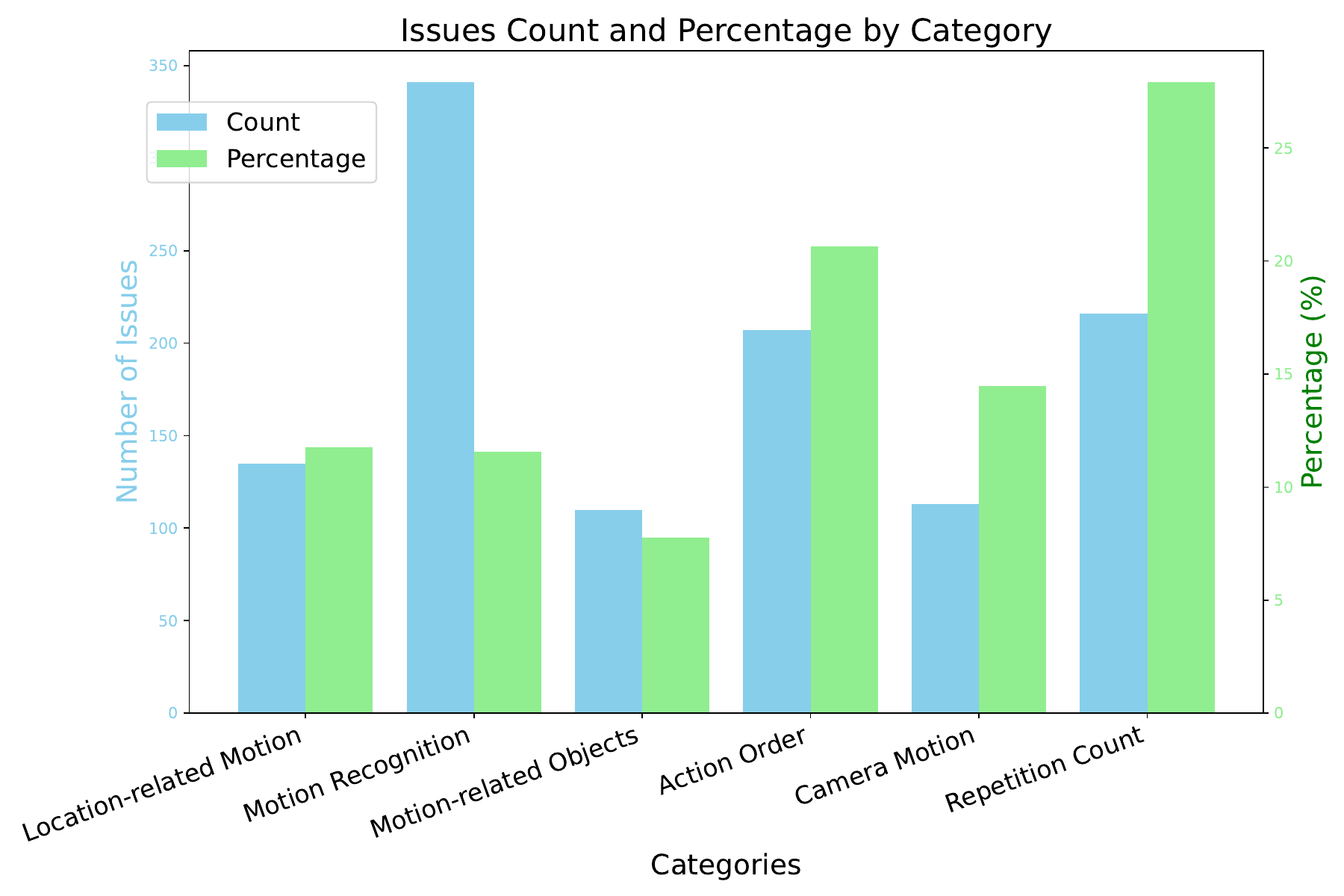}
    \caption{
    The absolute number and the proportion of questions that all models answered incorrectly relative to the total number of questions in each task type.
    }
    \vspace{-1em}
    \label{fig:allwrong_dist}
\end{figure}

\paragraph{Questions that confuses all models.}
As shown in \Tref{tab:dataset eval}, MotionBench is highly challenging for existing video understanding models. Currently, even the best video understanding models can achieve only less than 60\% accuracy. In MotionBench, there are some questions for which all models output incorrect answers. \Fref{fig:allwrong_dist} shows the absolute number and the proportion of questions that all models answered incorrectly relative to the total number of questions in each task type. Firstly, compared to the total number of questions in every task type, only a small fraction of questions were answered incorrectly by all models. Among the tasks, the highest proportion of questions that all models answered incorrectly is that in the ``Fast action count" task type. This attributes to counting repetitive actions at the motion level is inherently a very challenging task, and current video understanding models still struggle to handle such issues correctly.

\paragraph{Case study.} We show a case that all the models answered incorrectly. This is a case in which a male's hand is touching the car from the top and move to the lower left. However, most of the models believe that the video presents a hand ``tapping on the car surface". Such prediction is correct from a single image perspective, while in the video, the hand stays on the car surface and moves from the top to the lower left. Hence the gesture ``tapping" is not correct. This example demonstrates that single-frame predictions and perceptions can sometimes be misleading or even incorrect at the temporal level, which further underscores the value of creating a benchmark focused on motion-level temporal sequences.

\section{Limitations and Broader impact}
We propose MotionBench, a video understanding benchmark assessing the models' motion-level perception capability. However, there are several limitations to our approach that should be acknowledged. Firstly, although we have made efforts to include a diverse range of video content, our dataset may still have inherent biases in terms of geographical, cultural, and contextual variety. This could potentially limit the generalizability of research findings based on this dataset to different settings. Secondly, while we have performed extensive annotations, there may be occasional inaccuracies or inconsistencies due to human and automatic tool error.

Regarding the broader impact, motion-level perception is pivotal in video understanding. MotionBench provides a comprehensive benchmarking on video VLMs' motion-level perception. By making our dataset publicly available, we hope to further enhance the capabilities of video understanding models, thereby improving their applicability in real-world scenarios.

\section{More Dataset Samples}
For better demonstration, we show more samples from the MotionBench.

\begin{tcolorbox}[colback=blue!5!white, colframe=blue!40!white, title={uid=y26CvHFcz7BboSXN\_0}]
\begin{center}
    \includegraphics[width=1\linewidth]{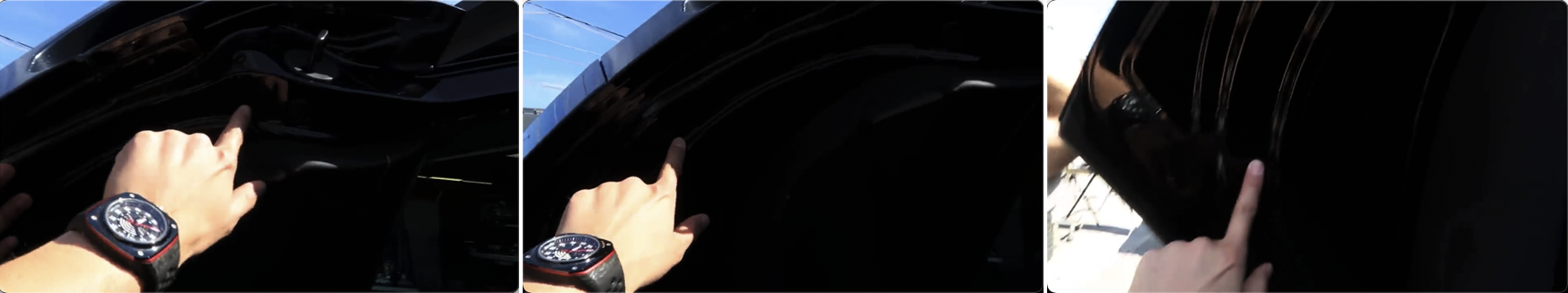}
\end{center}
What action does the hand in the video perform?

\textcolor{red!50!black}{A. Taps on the car surface (Gemini-1.5 pro, InternVL-40B, Oryx-34B, Qwen2-VL-72B)}

\textcolor{red!50!black}{B. Remains stationary (PLLaVA-34B)}

\textcolor{green!50!black}{C. Moves towards the lower left}

D. Waves back and forth

\end{tcolorbox}

\onecolumn
\begin{tcolorbox}[colback=blue!5!white, colframe=blue!40!white, title=Task type: Motion Recognition]
    \begin{center}
        \includegraphics[width=1\linewidth]{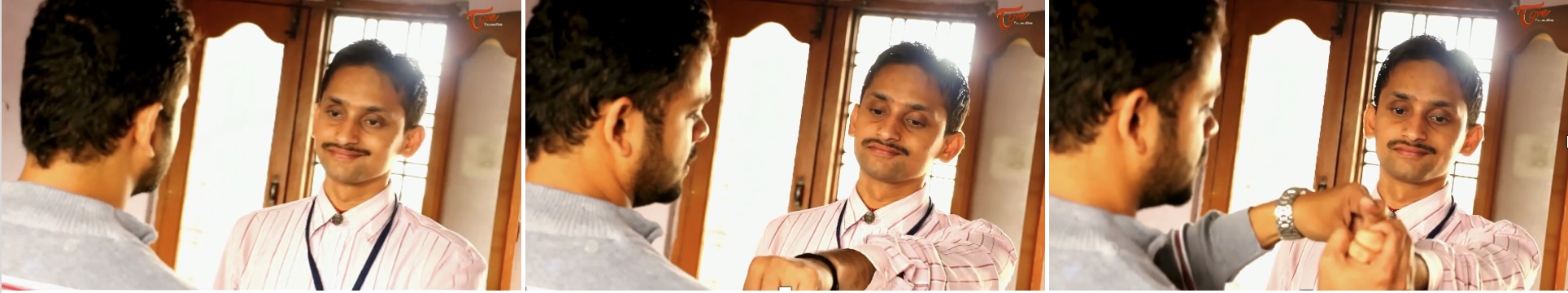}
    \end{center}
What is the sequence of movements between the two males?

\textcolor{green!50!black}{A. The male on the right raises his hand first, then the left male removes the string}

B. No movement occurs at all

C. Both actions occur simultaneously

D. The male on the left removes the string first, then the right male raises his hand
\end{tcolorbox}

\begin{tcolorbox}[colback=blue!5!white, colframe=blue!40!white, title=Task type: Motion Recognition]
    \begin{center}
        \includegraphics[width=1\linewidth]{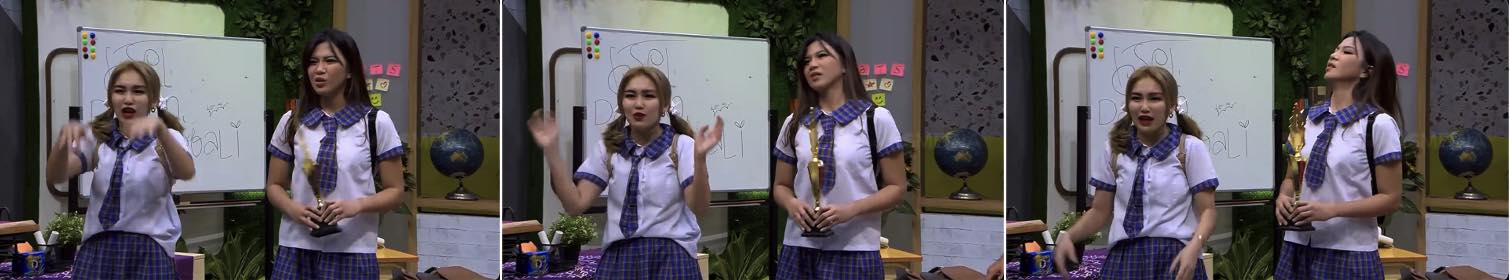}
    \end{center}

Which facial movement occurs with the woman on the right?

A. Full head tilt down

\textcolor{green!50!black}{B. Slight head turn to the right}

C. Eyes close briefly

D. Look straight ahead throughout
\end{tcolorbox}

\begin{tcolorbox}[colback=blue!5!white, colframe=blue!40!white, title=Task type: Action Order]
    \begin{center}
        \includegraphics[width=1\linewidth]{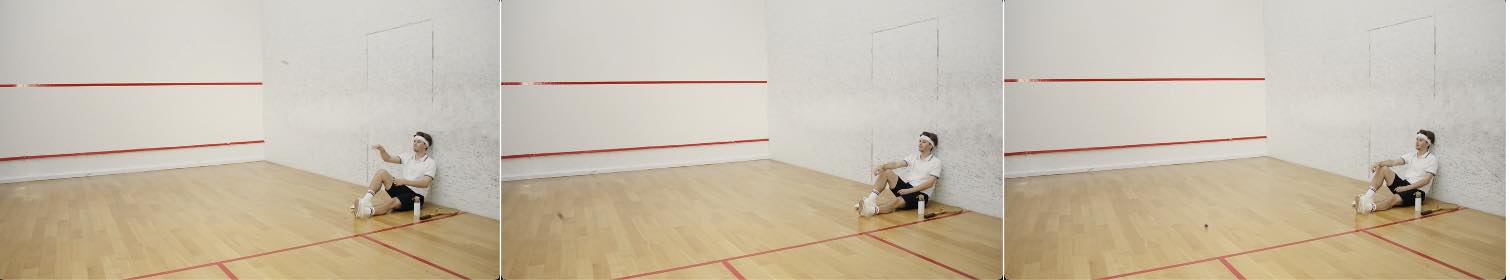}
    \end{center}

What is the sequence of ball movement in the video?

\textcolor{green!50!black}{A. The ball is thrown to the left and then rolls back from the left.}

B. The ball rolls from the left and then is thrown to the right.

C. The ball is thrown to the right and rolls from the right.

D. The ball is thrown upwards and rolls down.
\end{tcolorbox}

\begin{tcolorbox}[colback=blue!5!white, colframe=blue!40!white, title=Task type: Action Order]
    \begin{center}
        \includegraphics[width=1\linewidth]{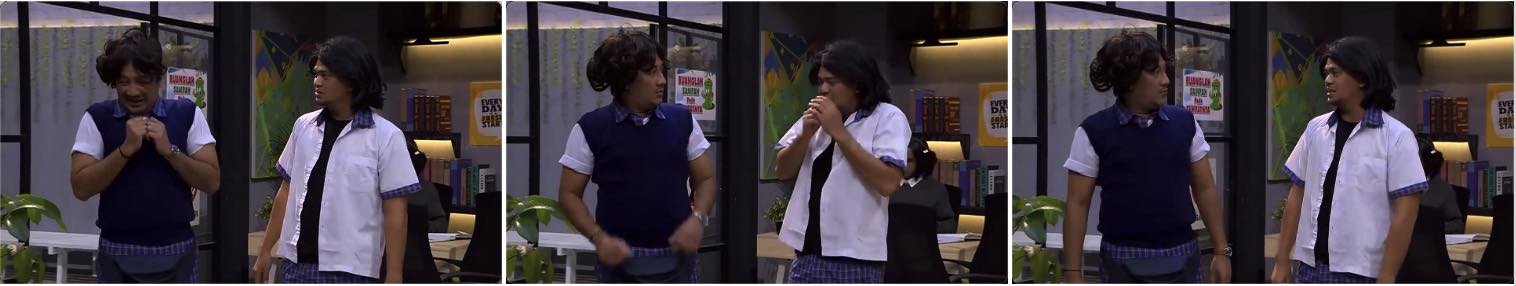}
    \end{center}

What is the sequence of actions involving the two men?

A. The man on the right raises his hands first, followed by the man on the left

B. Neither man raises their hands

\textcolor{green!50!black}{C. The man on the left raises his hands first, followed by the man on the right}

D. Both men raise their hands simultaneously
\end{tcolorbox}

\begin{tcolorbox}[colback=blue!5!white, colframe=blue!40!white, title=Task type: Motion-related Objects]
    \begin{center}
        \includegraphics[width=1\linewidth]{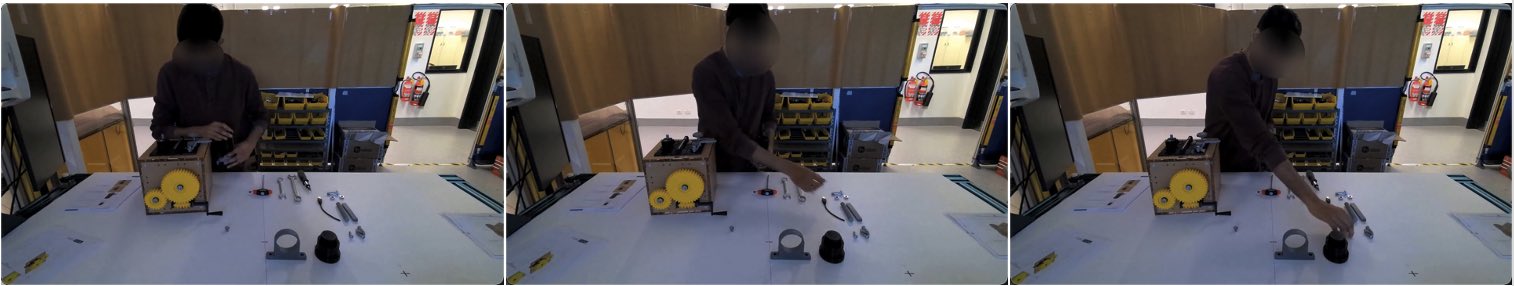}
    \end{center}

What did this person take with his right hand?

\textcolor{green!50!black}{A. Screw}

B. Thumbtack

C. Bolt nut

D. Pen core
\end{tcolorbox}

\begin{tcolorbox}[colback=blue!5!white, colframe=blue!40!white, title=Task type: Motion-related Objects]
    \begin{center}
        \includegraphics[width=1\linewidth]{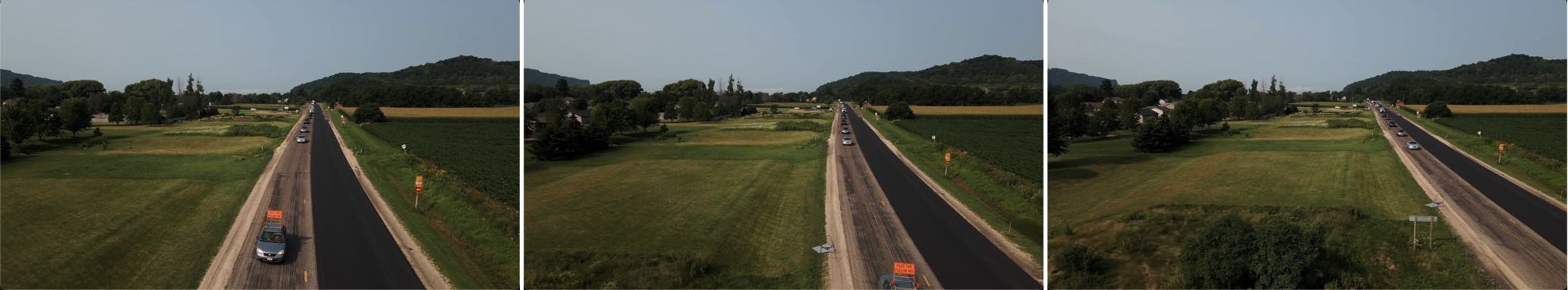}
    \end{center}

What does the camera reveal as it moves backward over the road?

A. A crosswalk appearing

\textcolor{green!50!black}{B. A sign on top of the lead car}

C. The end of a line of parked cars

D. The cars stopping abruptly
\end{tcolorbox}

\begin{tcolorbox}[colback=blue!5!white, colframe=blue!40!white, title=Task type: Location-related Motion]
    \begin{center}
        \includegraphics[width=1\linewidth]{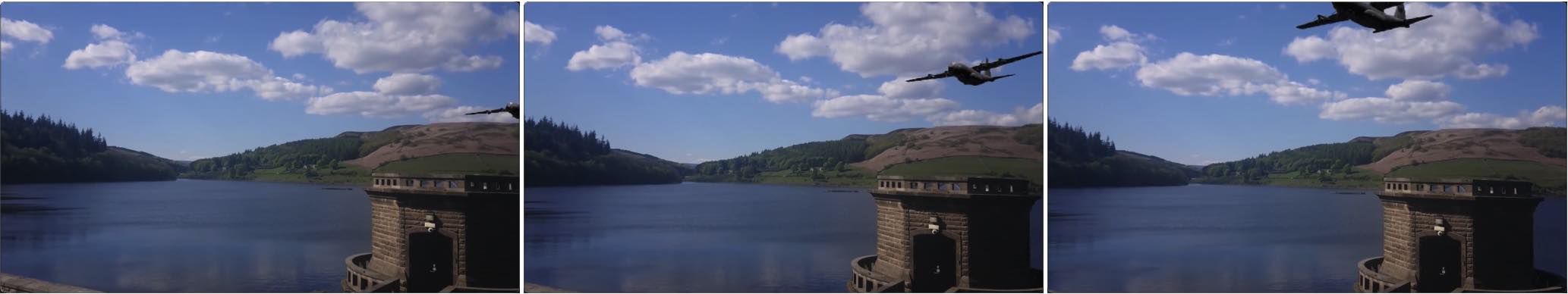}
    \end{center}

In what order does the plane appear and move across the screen?

A. From the left to completely leaving the frame

B. From the top left to bottom right

\textcolor{green!50!black}{C. From the right to the upper part before disappearing}

D. From the bottom left to top right
\end{tcolorbox}

\begin{tcolorbox}[colback=blue!5!white, colframe=blue!40!white, title=Task type: Location-related Motion]
    \begin{center}
        \includegraphics[width=1\linewidth]{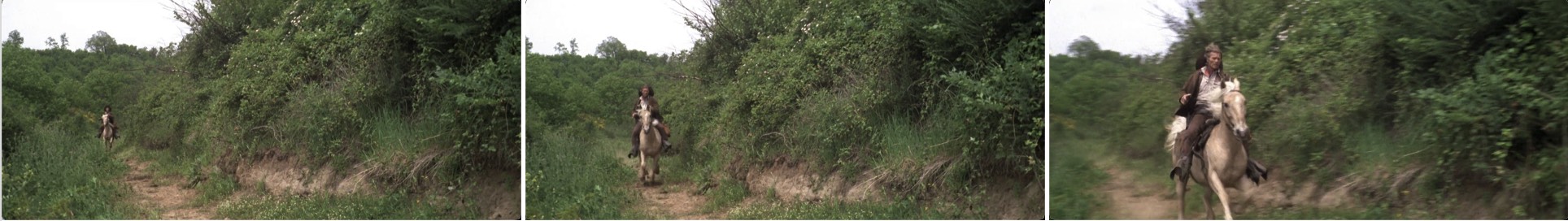}
    \end{center}
What movement trajectory does the horse follow?

A. Stays in place

\textcolor{green!50!black}{B. Moves directly towards the camera}

C. Gallops in circles

D. Jumps over an obstacle
\end{tcolorbox}

\begin{tcolorbox}[colback=blue!5!white, colframe=blue!40!white, title=Task type: Repetition Count]
    \begin{center}
        \includegraphics[width=1\linewidth]{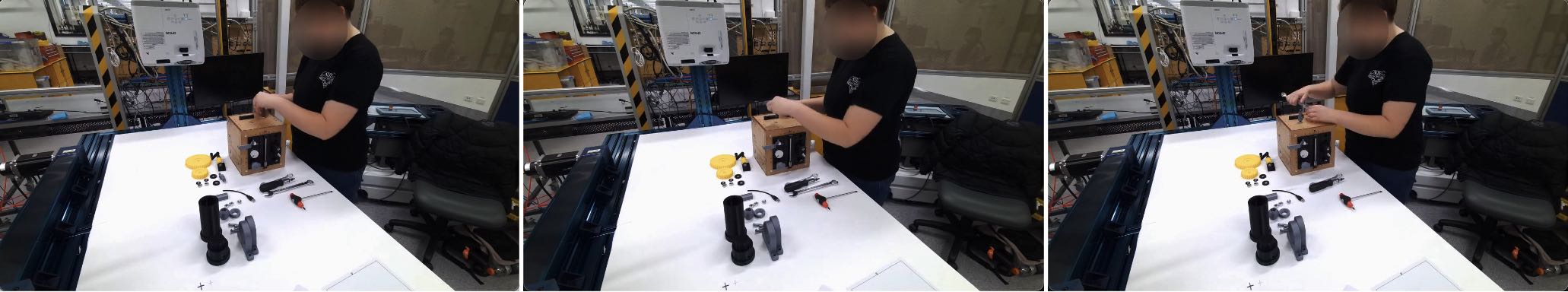}
    \end{center}

Please count the number of repeated actions in the video.

\textcolor{green!50!black}{A. 3}

B. 6

C. 9

D. 4

\end{tcolorbox}

\begin{tcolorbox}[colback=blue!5!white, colframe=blue!40!white, title=Task type: Repetition Count]
    \begin{center}
        \includegraphics[width=0.6\linewidth]{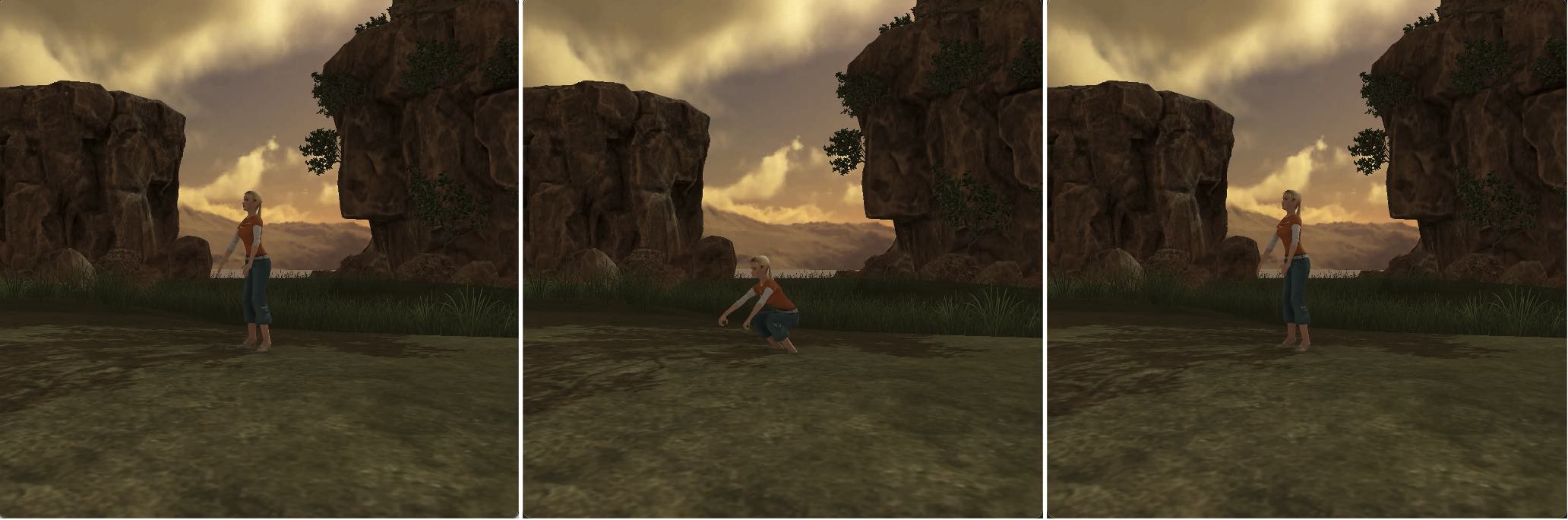}
    \end{center}

Please count the number of repeated actions in the video.

A. 3

\textcolor{green!50!black}{B. 2}

C. 1

D. 6
\end{tcolorbox}

\begin{tcolorbox}[colback=blue!5!white, colframe=blue!40!white, title=Task type: Camera Motion]
    \begin{center}
        \includegraphics[width=1\linewidth]{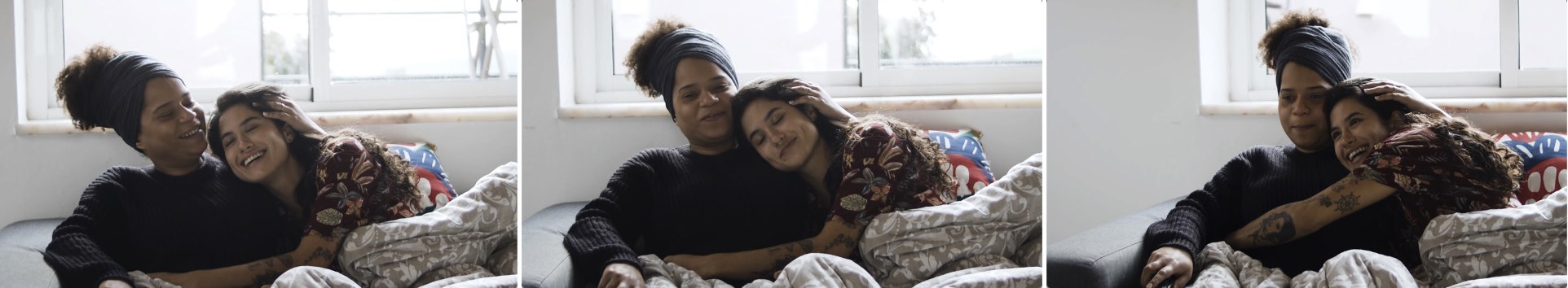}
    \end{center}
Does the camera perform any movement during the scene?

A. Yes, it zooms in.

B. No, it remains static.

C. Yes, it pans to the left.

\textcolor{green!50!black}{D. Yes, it rotates counterclockwise.}
\end{tcolorbox}

\begin{tcolorbox}[colback=blue!5!white, colframe=blue!40!white, title=Task type: Camera Motion]
    \begin{center}
        \includegraphics[width=1\linewidth]{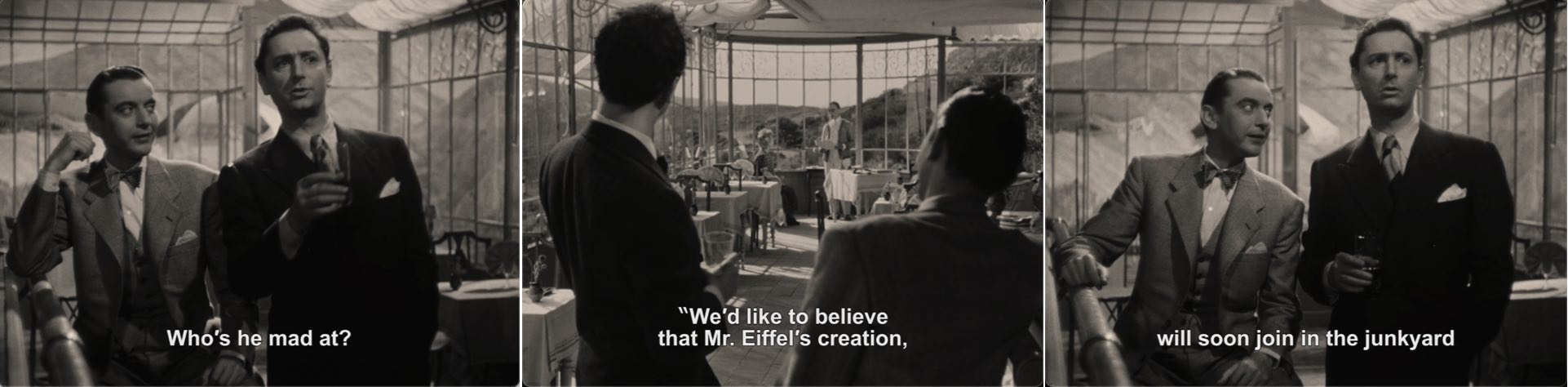}
    \end{center}

What is the sequence of the camera movements during the interaction?

A. The camera stays still, only focusing on the woman

B. The camera shifts to show the men with their backs, then returns to face the men

C. The camera shifts to show both men together, then moves back to the woman

\textcolor{green!50!black}{D. The camera starts facing the men, shifts to the woman, then moves back to the men}
\end{tcolorbox}

\twocolumn




\end{document}